\documentclass[10pt,twocolumn,letterpaper]{article}

\usepackage{cvpr}              

\usepackage{graphicx}
\usepackage{amsmath}
\usepackage{amssymb}
\usepackage{booktabs}
\usepackage[accsupp]{axessibility}
\usepackage{times}
\usepackage{epsfig}
\usepackage{multirow}
\usepackage{multicol}
\usepackage{bm}
 
\makeatletter
\newcommand{\thickhline}{%
    \noalign {\ifnum 0=`}\fi \hrule height 1pt
    \futurelet \reserved@a \@xhline
}
\usepackage[colorlinks,linkcolor=red]{hyperref}

\usepackage[capitalize]{cleveref}
\crefname{section}{Sec.}{Secs.}
\Crefname{section}{Section}{Sections}
\Crefname{table}{Table}{Tables}
\crefname{table}{Tab.}{Tabs.}

\def\cvprPaperID{4609} 
\def\confName{CVPR}
\def\confYear{2022}
\newcommand\ourmethod{ACmix}

\begin{document}

\title{On the Integration of Self-Attention and Convolution}

\author{%
  Xuran Pan$^{1}$\ \ \
  Chunjiang Ge$^{1}$\ \ \
  Rui Lu$^{1}$\ \ \
  Shiji Song$^{1}$\ \ \
  Guanfu Chen$^{2}$ \ \ \
  Zeyi Huang$^{2}$ \ \ \ 
  Gao Huang$^{1,3}$\thanks{Corresponding author.}\\
    $^{1}$Department of Automation, BNRist, Tsinghua University, Beijing, China\\
    $^{2}$Huawei Technologies Ltd., China\\
    $^{3}$Beijing Academy of Artificial Intelligence, Beijing, China\\
  \texttt{\small \{pxr18, gecj20, r-lu21\}@mails.tsinghua.edu.cn} \\ \texttt{\small{} \{chenguanfu1, huangzeyi2\}@huawei.com},
  \texttt{\small\{shijis, gaohuang\}@tsinghua.edu.cn}
}
\maketitle

\begin{abstract}
Convolution and self-attention are two powerful techniques for representation learning, and they are usually considered as two peer approaches that are distinct from each other. In this paper, we show that there exists a strong underlying relation between them, in the sense that the bulk of computations of these two paradigms are in fact done with the same operation. Specifically, we first show that a traditional convolution with kernel size $k\!\times \!k$ can be decomposed into $k^2$ individual $1\!\times \!1$ convolutions, followed by shift and summation operations. Then, we interpret the projections of queries, keys, and values in self-attention module as multiple $1\!\times \!1$ convolutions, followed by the computation of attention weights and aggregation of the values. Therefore, the first stage of both two modules comprises the similar operation. More importantly, the first stage contributes a dominant computation complexity (square of the channel size) comparing to the second stage. This observation naturally leads to an elegant integration of these two seemingly distinct paradigms, i.e., a \textbf{mixed} model that enjoys the benefit of both self-\textbf{A}ttention and \textbf{C}onvolution (\ourmethod), while having minimum computational overhead compared to the pure convolution or self-attention counterpart. Extensive experiments show that our model achieves consistently improved results over competitive baselines on image recognition and downstream tasks. Code and pre-trained models will be released at \url{https://github.com/LeapLabTHU/ACmix} and \url{https://gitee.com/mindspore/models}.
\end{abstract}


\section{Introduction}
\begin{figure}
    \begin{center}
    \includegraphics[width=0.9\linewidth]{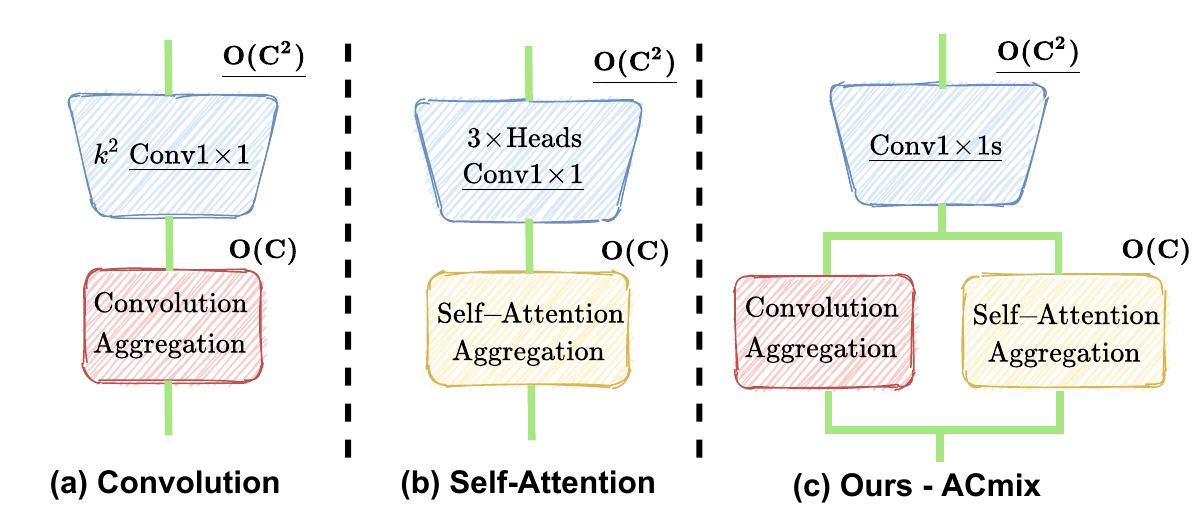}
    \vskip -0.1in
    \caption{A sketch of \ourmethod. We explore a closer relationship between convolution and self-attention in the sense of sharing the same computation overhead ($1\!\times \!1$ convolutions), and combining with the remaining lightweight aggregation operations. 
    We show the computation complexity of each block \textit{w.r.t} the feature channel.}
    \label{fig:1}
    \end{center}
    \vskip -0.3in
\end{figure}
Recent years have witnessed the vast development of convolution and self-attention in computer vision. Convolution neural networks (CNNs) are widely adopted on image recognition \cite{he2016deep,huang2019convolutional}, semantic segmentation \cite{chen2017deeplab} and object detection \cite{ren2015faster}, and achieve state-of-the-art performances on various benchmarks. On the other hand, self-attention is first introduced in natural language processing \cite{bahdanau2014neural, vaswani2017attention}, and also shows great potential in the fields of image generation and super-resolution \cite{parmar2018image,child2019generating}. More recently, with the advent of vision transformers \cite{dosovitskiy2020image,ramachandran2019stand,carion2020end}, attention-based modules have achieved comparable or even better performances than their CNN counterparts on many vision tasks.

Despite the great success that both approaches have achieved, convolution and self-attention modules usually follow different design paradigms. Traditional convolution leverages an aggregation function over a localized receptive field according to the convolution filter weights, which are shared in the whole feature map. The intrinsic characteristics impose crucial inductive biases for image processing. Comparably, the self-attention module applies a weighted average operation based on the context of input features, where the attention weights are computed dynamically via a similarity function between related pixel pairs. The flexibility enables the attention module to focus on different regions adaptively and capture more informative features.

Considering the different and complementary properties of convolution and self-attention, there exists a potential possibility to benefit from both paradigms by integrating these modules. Previous work has explored the combination of self-attention and convolution from several different perspectives. Researches from early stages, e.g., SENet \cite{hu2018squeeze}, CBAM \cite{woo2018cbam}, show that self-attention mechanism can serve as an augmentation for convolution modules. More recently, self-attention modules are proposed as individual blocks to substitute traditional convolutions in CNN models, e.g., SAN \cite{zhao2020exploring}, BoTNet \cite{srinivas2021bottleneck}. Another line of research focuses on combining self-attention and convolution in a single block, e.g., AA-ResNet \cite{bello2019attention}, Container \cite{gao2021container}, while the architecture is limited in designing independent paths for each module. Therefore, existing approaches still treat self-attention and convolution as distinct parts, and the underlying relations between them have not been fully exploited.

In this paper, we seek to unearth a closer relationship between self-attention and convolution. By decomposing the operations of these two modules, we show that they heavily rely on the same $1\!\times \!1$ convolution operations. Based on this observation, we develop a mixed model, named \ourmethod, and integrate self-attention and convolution elegantly with minimum computational overhead. Specifically, we first project the input feature maps with $1\!\times \!1$ convolutions and obtain a rich set of intermediate features. Then, the intermediate features are reused and aggregated following different paradigms, i.e, in self-attention and convolution manners respectively. In this way, \ourmethod \ enjoys the benefit of both modules, and effectively avoids conducting expensive projection operations twice.

To summarize, our contributions are two folds:

(1) A strong underlying relation between self-attention and convolution is revealed, providing new perspectives on understanding the connections between two modules and inspirations for designing new learning paradigms.

(2) An elegant integration of the self-attention and convolution module, which enjoys the benefits of both worlds, is presented. Empirical evidence demonstrates that the hybrid model outperforms its pure convolution or self-attention counterpart consistently.
\section{Related Work}
Convolution neural networks \cite{LeCun1989BackpropagationAT,Krizhevsky2012ImageNetCW}, which use convolution kernels to extract local features, have become the most powerful and conventional technique for various vision tasks \cite{Simonyan2015VeryDC,he2016deep,huang2017densely}. Meanwhile, self-attention also demonstrated its prevailing performance on a broad range of language tasks like BERT and GPT3 \cite{devlin2018bert,radford2018improving,NEURIPS2020_1457c0d6}. Theoretical analysis \cite{cordonnier2019relationship} indicates that, when equipped with sufficiently large capacity, self-attention can express the function class of any convolution layers. Therefore, a line of research recently explores the possibility of adopting the self-attention mechanism into vision tasks \cite{dosovitskiy2020image,hu2018squeeze}. There are two mainstream methods, one uses self-attention as building blocks in a network \cite{carion2020end,zheng2020rethinking,pan20203d}, and another views self-attention and convolution as complementary parts \cite{wang2018non,Li2019SelectiveKN,Cao2019GCNetNN}.

\subsection{Self-Attention only}
Inspired by the power of self-attention's expressive ability in long-range dependencies \cite{vaswani2017attention,devlin2018bert}, a march of work endeavours to solely use self-attention as elementary building blocks to construct the model for vision tasks \cite{dosovitskiy2020image,carion2020end,Beal2020TowardTO}. Some works \cite{ramachandran2019stand,zhao2020exploring} show that self-attention can become a stand-alone primitive for vision models which completely substitute convolutional operations. Recently, Vision Transformer \cite{dosovitskiy2020image} shows that given enough data, we can treat an image as a sequence of 256 tokens and leverage Transformer models \cite{vaswani2017attention} to achieve competitive results in image recognition. Furthermore, transformer paradigm is adopted in detection \cite{carion2020end,Beal2020TowardTO,Zhu2020DeformableDD}, segmentation \cite{zheng2020rethinking,Zhang2020FeaturePT,Wang2020EndtoEndVI}, point cloud recognition \cite{pan20203d,Guo2020PCTPC} and other vision tasks \cite{parmar2018image,Chen2020PreTrainedIP}.

\subsection{Attention enhanced Convolution}
Multiple previously proposed attention mechanisms over images suggest it can overcome the limitation of locality for convolutional networks. Therefore, many researchers explore the possibility of employing attention modules or utilizing more relational information to enhance the functionality of convolutional networks. Particularly, Squeeze-and-Excitation (SE) \cite{hu2018squeeze} and Gather-Excite (GE) \cite{hu2018gather} reweigh the map for each channel. BAM \cite{park2018bam} and CBAM \cite{woo2018cbam} independently reweigh both channels and spatial locations to better refine the feature map. AA-Resnet\cite{bello2019attention} augments certain convolutional layers by concatenating attention maps from another independent self-attention pipeline. BoTNet \cite{srinivas2021bottleneck} substitutes convolutions with self-attention modules at late stages of the model. Some work aims at designing a more flexible feature extractor by aggregating information from a wider range of pixels. Hu \etal \cite{hu2019local} proposed a local-relation approach to adaptively determine aggregation weights based on the compositional relationship of local pixels. Wang \etal proposed non-local network \cite{wang2018non}, which increases the receptive field by introducing non-local blocks that compare similarity among global pixels.

\begin{figure*}
    \begin{center}
    \includegraphics[width=0.9\linewidth]{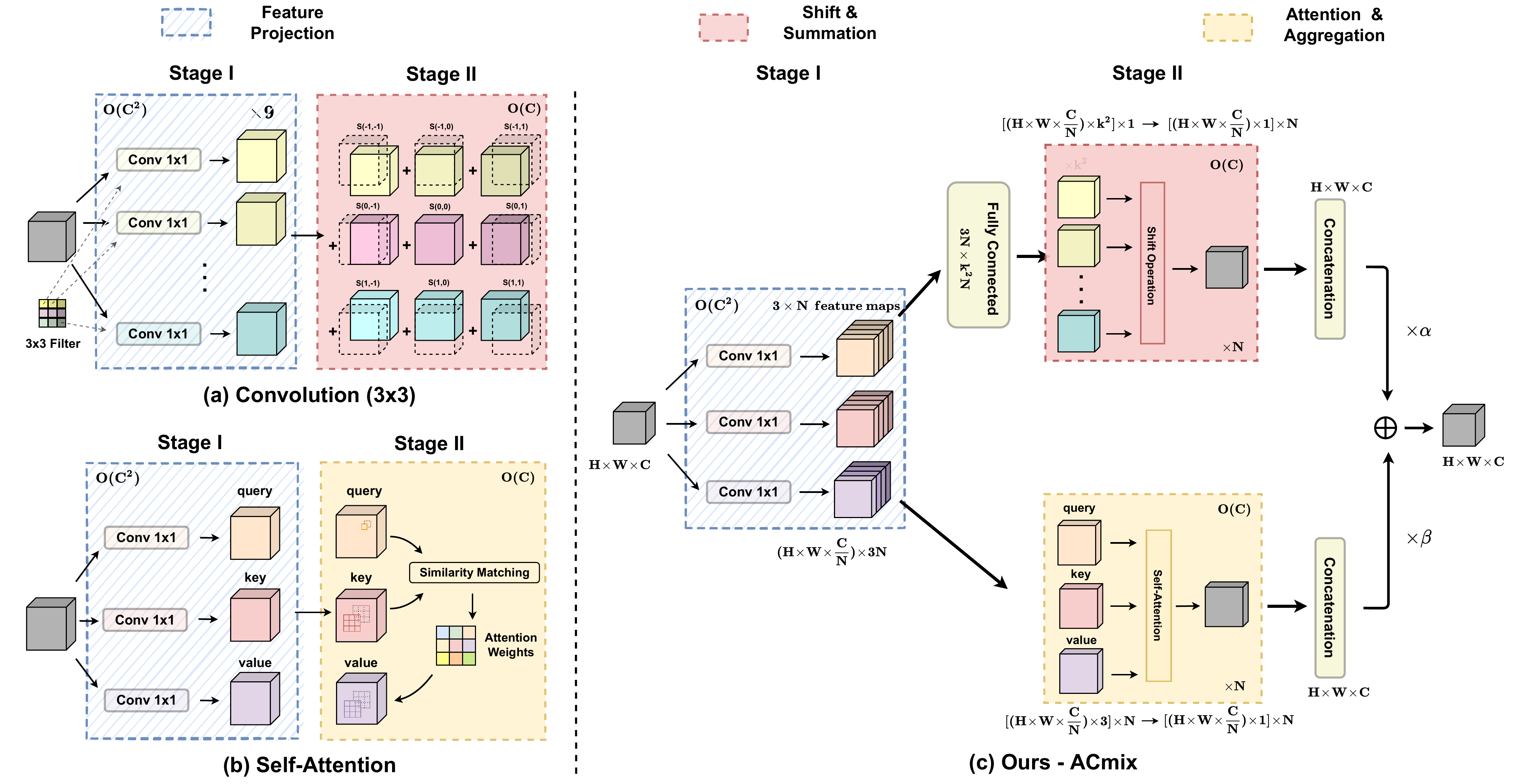}
    \vskip -0.1in
    \caption{An illustration of the proposed hybrid module. The left figures show the pipeline of traditional convolution and self-attention modules. (a) Convolution. The output of a $3\!\times \!3$ convolution can be decomposed as a summation of shifted feature maps, where each feature map is obtained by performing a $1\!\times \!1$ convolution concerning the kernel weights from a certain position. $s(x,y)$ corresponds to the Shift operation defined in Sec.\ref{re_conv}. (b) Self-Attention. Input feature map is first projected as queries, keys, and values with $1\!\times \!1$ convolutions. The attention weights computed by queries and keys are adopted to aggregate the values. The right figure shows the pipeline of our module. (c) \ourmethod. At stage I, the input feature map is projected with three $1\!\times \!1$ convolutions. At stage II, the intermediate features are used following two paradigms respectively. Features from both paths are added together and serve as the final output.
    The computational complexity of each operation block is marked at the upper corner.}
    \label{fig:2}
    \end{center}
    \vskip -0.3in
\end{figure*}

\subsection{Convolution enhanced Attention}
With the advent of Vision Transformer \cite{dosovitskiy2020image}, numerous transformer-based variants have been proposed and achieved significant improvements on computer vision tasks. Among which exist researches focusing on complementing transformer models with convolution operations to introduce additional inductive biases. CvT \cite{wu2021cvt} adopts convolution in the tokenization process and utilize stride convolution to reduce computation complexity of self-attention. ViT with convolutional stem \cite{xiao2021early} proposes to add convolutions at the early stage to achieve stabler training. CSwin Transformer \cite{dong2021cswin} adopts a convolution-based positional encoding technique and shows improvements on downstream tasks. Conformer \cite{peng2021conformer} combines Transformer with an independent CNN model to integrate both features. 

\section{Revisiting Convolution and Self-Attention}
\label{revisit}
Convolution and self-attention have been widely known in their current forms. To better capture the relationship between these two modules, we revisit them from a novel view by decomposing their operations into separated stages.

\subsection{Convolution}
\label{re_conv}
Convolution is one of the most essential parts of modern ConvNets. We first review the standard convolution operation and reformulate it from a different perspective. The illustration is shown in Fig.\ref{fig:2}(a). For simplicity, we assume the stride of convolution is 1.

Consider a standard convolution with the kernel $K \in \mathcal{R}^{C_{\rm{out}} \times C_{\rm{in}} \times k \times  k}$, where $k$ is the kernel size and $C_{\rm{in}},C_{\rm{out}}$ are the input and output channel size. Given tensors $F \in \mathcal{R}^{C_{\rm{in}}\times H\times W}, G \in \mathcal{R}^{C_{\rm{out}}\times H\times W}$ as the input and output feature maps, where $H, W$ denote the height and width, we denote $f_{ij} \in \mathcal{R}^{C_{\rm{in}}},g_{ij} \in \mathcal{R}^{C_{\rm{out}}}$ as the feature tensors of pixel $(i,j)$ corresponding to $F$ and $G$ respectively.
Then, the standard convolution can be formulated as:
\begin{equation}
\setlength{\abovedisplayskip}{1ex}
    \label{conv1}
    g_{ij} = \sum_{p, q} K_{p, q}f_{i+p-\lfloor k/2 \rfloor, j+q-\lfloor k/2 \rfloor},
\setlength{\belowdisplayskip}{1ex}
\end{equation}
where $K_{p, q} \in \mathcal{R}^{C_{\rm{out}} \times C_{\rm{in}}}, \ p,q\in\{0,1,\dots,k\!-\!1\},$ represents the kernel weights with regard to the indices of the kernel position $(p,q)$.

For convenience, we can rewrite Eq.(\ref{conv1}) as the summation of the feature maps from different kernel positions:
\begin{equation}
\setlength{\abovedisplayskip}{1ex}
\label{conv_stage1}
    g_{ij} = \sum_{p,q}g_{ij}^{(p,q)},
\setlength{\belowdisplayskip}{1ex}
\end{equation}
with
\begin{equation}
\setlength{\abovedisplayskip}{1ex}
\label{conv_stage2}
    g_{ij}^{(p,q)}= K_{p, q}f_{i+p-\lfloor k/2 \rfloor, j+q-\lfloor k/2 \rfloor}.
\setlength{\belowdisplayskip}{1ex}
\end{equation}
To further simplify the formulation, we define the \textbf{Shift} operation, $\tilde{f} \triangleq {\rm Shift}(f, \Delta x, \Delta y)$, as:
\begin{equation}
\setlength{\abovedisplayskip}{1ex}
    \tilde{f}_{i,j} = f_{i+\Delta x, j+\Delta y}, \ \forall i,j,
\setlength{\belowdisplayskip}{1ex}
\end{equation}
where $\Delta x, \Delta y$ correspond to the horizontal and vertical displacements. 
Then, Eq.(\ref{conv_stage2}) can be rewritten as:
\begin{align}
    g_{ij}^{(p,q)}&= K_{p, q}f_{i+p-\lfloor k/2 \rfloor, j+q-\lfloor k/2 \rfloor} \nonumber\\
    &={\rm Shift}(K_{p, q}f_{ij}, p-\lfloor k/2 \rfloor, q-\lfloor k/2 \rfloor).
\end{align}
As a result, the standard convolution can be summarized as two stages:
\begin{align}
    &{\rm \textbf{Stage \ I:}} \ \ \ \tilde{g}_{ij}^{(p,q)} = K_{p, q}f_{ij}, \\
    \label{conv2}
    &{\rm \textbf{Stage \ II:}} \ g^{(p,q)}_{ij}\! =\! {\rm Shift}(\tilde{g}^{(p,q)}_{ij}, \!p\!-\!\lfloor k/2 \rfloor,\! q\!-\!\lfloor k/2 \rfloor),\\
    \label{conv3}
    &\quad \quad \quad \quad \ g_{ij} = \sum_{p,q}g_{ij}^{(p,q)}.
\end{align}

At the first stage, the input feature map is linearly projected \textit{w.r.t.} the kernel weights from a certain position, i.e., ($p,q$). This is the same as a standard $1\!\times \!1$ convolution. While in the second stage, the projected feature maps are shifted according to the kernel positions and finally aggregated together. 
It can be easily observed that most of the computational costs are performed in the $1\!\times \!1$ convolution, while the following shift and aggregation are lightweight.

\begin{table}[t]
    \begin{center}
    \small
    \setlength{\tabcolsep}{0.1mm}{
    \renewcommand\arraystretch{0.9}
    \begin{tabular}{cc|c|c|c|c}
    \toprule
     & & \multicolumn{2}{c|}{\textbf{$\quad$ Theoretical}} & \multicolumn{2}{c}{\textbf{ResNet 50}}\\
    \textbf{Module} & \textbf{Stg} & \textbf{FLOPs($\!\times{}\!\bf{hw}$)} & \textbf{Params} & \textbf{FLOPs(G)} & \textbf{Params(M)}\\
    \midrule
    \multirow{2}{*}{Conv} & I & $k_c^2 \ C^2$ & $k_c^2 \ C^2$ & $1.9(99\%)$ & $11.3(100\%)$\\
    & II & $k_c^2 \ C$ & 0 & $0.1(1\%)$ & $0(0\%)$\\
    \midrule
    Self & I & $3C^2$ & $3C^2$ & $1.0(83\%)$ & $3.8(100\%)$\\
    Attention & II & $2k_{a}^2 \ C$ & 0 & $0.2(17\%)$ & $0(0\%)$\\
    \midrule
    \specialrule{0em}{-0pt}{-3pt}
    \multirow{4}{*}{\textbf{\ourmethod}} & \multirow{2}{*}{I} & \multirow{2}{*}{$3C^2$} & \multirow{2}{*}{$3C^2$} & \multirow{2}{*}{$1.0(73\%)$} & \multirow{2}{*}{$3.8(92\%)$}\\
    & & & & \\
    \specialrule{0em}{-3pt}{-1pt}
    & \multirow{2}{*}{II} & $(k_c^2\!+\!2k_{a}^2)C$ & \multirow{2}{*}{$\textcolor{red}{3k_c^2N\!+\!k_c^4C}$} & \multirow{2}{*}{$0.4(27\%)$} & \multirow{2}{*}{$0.3(8\%)$}\\
    \specialrule{0em}{-1pt}{-1pt}
    & & $\textcolor{red}{+\!(3k_{c}^2\!+\!k_c^4)C}$ & & \\
    \bottomrule
    \end{tabular}}
    \end{center}
    \vskip -0.2in
    \caption{FLOPs and Parameters for different modules at two stages. $\mathbf{C}$: Input and output channel. $\mathbf{h,w}$: Length and width of feature map. $\mathbf{k_c}$: Kernel size for convolution. $\mathbf{k_a}$: Kernel size for self-attention. $\mathbf{N}$: Head of self-attention. Numbers in red correspond to the additional FLOPs/Params introduced by \ourmethod. Percentages within the brackets are fractions of the whole module.}
    \label{flops}
    \vskip -0.2in
\end{table}

\subsection{Self-Attention}
Attention mechanism has also been widely adopted in vision tasks. Comparing to the traditional convolution, attention allows the model to focus on important regions within a larger size context.
We show the illustration in Fig.\ref{fig:2}(b).

Consider a standard self-attention module with $N$ heads. Let $F \in \mathcal{R}^{C_{\rm{in}}\times H\times W}, G \in \mathcal{R}^{C_{\rm{out}}\times H\times W}$ denote the input and output feature. Let $f_{ij} \in \mathcal{R}^{C_{\rm{in}}},g_{ij} \in \mathcal{R}^{C_{\rm{out}}}$ denote the corresponding tensor of pixel $(i,j)$. Then, output of the attention module is computed as:
\begin{equation}
\setlength{\abovedisplayskip}{1ex}
    \label{sa1}
    g_{ij}\! =\! \overset{N}{\underset{l=1}{||}}\!\left(\sum_{a,b\in \mathcal{N}_k(i,j)}\!\!\!\!\!\!\!{\rm A}(W_q^{(l)}f_{ij}, W_k^{(l)}f_{ab})W_v^{(l)}f_{ab}\right)\!,
\setlength{\belowdisplayskip}{1ex}
\end{equation}
where $||$ is the concatenation of the outputs of $N$ attention heads, and $W_q^{(l)}, W_k^{(l)}, W_v^{(l)}$ are the projection matrices for queries, keys and values. $\mathcal{N}_k(i,j)$ represents a local region of pixels with spatial extent $k$ centered around $(i,j)$, and ${\rm A}(W_q^{(l)}f_{ij}, W_k^{(l)}f_{ab})$ is the corresponding attention weight with regard to the features within $\mathcal{N}_k(i,j)$.

For the widely adopted self-attention modules in \cite{hu2019local,ramachandran2019stand}, the attention weights are computed as:
\begin{equation}
\label{pairwise}
    {\rm A}(W_q^{(l)}\!f_{ij},\! W_k^{(l)}\!f_{ab})\! =\! {\underset{\mathcal{N}_k(i,j)}{\rm softmax}}\!\left(\frac{\!(W_q^{(l)}\!f_{ij})\!^{\rm T}\! (W_k^{(l)}\!f_{ab})}{\sqrt{d}}\!\right)\!,
\end{equation}
where $d$ is the feature dimension of $W_q^{(l)}f_{ij}$.

Also, multi-head self-attention can be decomposed into two stages, and reformulated as:
\begin{align}
    &{\rm \textbf{Stage \ I:}} \ q_{ij}^{(l)}\!\! =\! W_q^{(l)}\!f_{ij}, \!k_{ij}^{(l)}\! =\! W_k^{(l)}\!f_{ij},\! v_{ij}^{(l)}\! =\! W_v^{(l)}\!f_{ij},\\
    \label{sa3}
    &{\rm \textbf{Stage \ II:}} \ g_{ij}\! =\! \overset{N}{\underset{l=1}{||}}\!\left(\sum_{a,b\in \mathcal{N}_k(i,j)}\!\!\!\!\!\!\!{\rm A}(q_{ij}^{(l)}, k_{ab}^{(l)})v_{ab}^{(l)}\right)\!.
\end{align}

Similar to the traditional convolution in Sec.\ref{re_conv}, $1\!\times \!1$ convolutions are first conducted in stage I to project the input feature as query, key and value. On the other hand, Stage II comprises the calculation of the attention weights and aggregation of the value matrices, which refers to gathering local features. The corresponding computational cost is also proved to be minor comparing to Stage I, following the same pattern as convolution.

\begin{figure*}
    \begin{center}
    \includegraphics[width=0.9\linewidth]{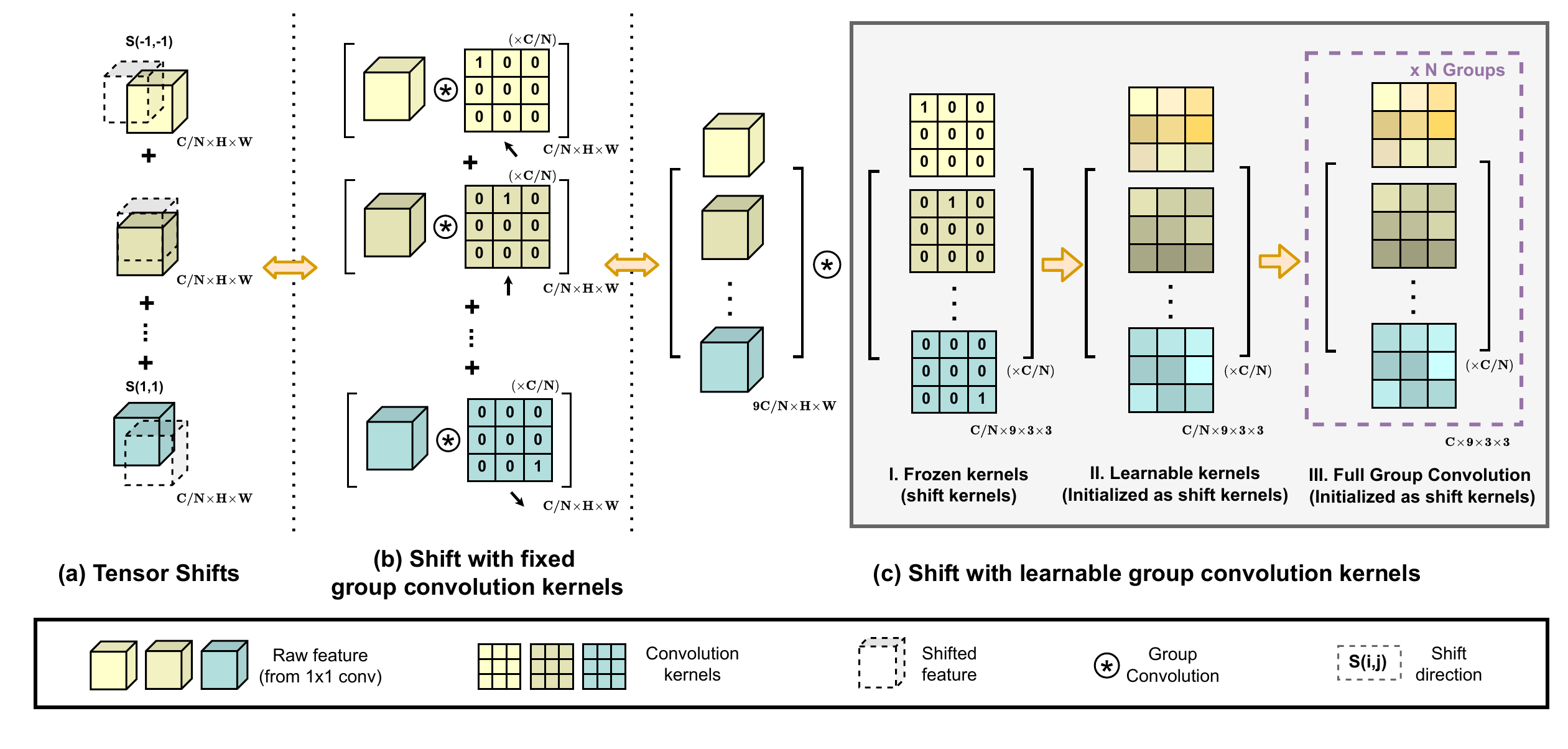}
    \vskip -0.2in
    \caption{Practical improvements on shift operations. (a) Simple implementation with tensor shifts. (b) Fast implementation with carefully designed group convolution kernels. (c) Further adaptations with learnable kernels and multiple convolution groups.}
    \label{fig:3}
    \end{center}
    \vskip -0.3in
\end{figure*}

\subsection{Computational Cost}
To fully understand the computation bottleneck of the convolution and self-attention modules, we analyse the floating-point operations (FLOPs) and the number of parameters at each stage and summarize in Tab.\ref{flops}. It is shown that theoretical FLOPs and parameters at Stage I of convolution have quadratic complexity with regard to the channel size $C$, while the computational cost for Stage II is linear to $C$ and no additional training parameters are required. 

A similar trend is also found for the self-attention module, where all training parameters are preserved at Stage I. As for the theoretical FLOPs, we consider a normal case in a ResNet-like model where $k_a\!=\!7$ and $C\!=\!64,128,256,512$ for various layer depths. It is explicitly shown that Stage I consumes a heavier operation as $3C^2>2k_{a}^2C$, and the discrepancy is more distinct as channel size grows.

To further verify the validity of our analysis, we also summarize the actual computational costs of the convolution and self-attention modules in a ResNet50 model in Tab.\ref{flops}. 
We practically add up the costs of all $3\!\times \!3$ convolution (or self-attention) modules to reflect the tendency from the model perspective.
It is shown that 99\% computation of convolution and 83\% of self-attention are conducted at Stage I, which are consistent with our theoretical analysis.

\section{Method}

\subsection{Relating Self-Attention with Convolution}
\label{revisit3}

The decomposition of self-attention and convolution modules in Sec.\ref{revisit} has revealed deeper relations from various perspectives. First, the two stages play quite similar roles. Stage I is a feature learning module, where both approaches share the same operations by performing $1\!\times \!1$ convolutions to project features into deeper spaces. On the other hand, stage II corresponds to the procedure of feature aggregation, despite the differences in their learning paradigms. 

From the computation perspective, the $1\!\times \!1$ convolutions conducted at Stage I of both convolution and self-attention modules require a quadratic complexity of theoretical FLOPs and parameters with regard to the channel size $C$. Comparably, at stage II both modules are lightweight or nearly free of computation. 

As a conclusion, the above analysis shows that (1) Convolution and self-attention practically share the same operation on projecting the input feature maps through $1\!\times \!1$ convolutions, which is also the computation overhead for both modules.\! (2) Although crucial for capturing semantic features, the aggregation operations at stage II are lightweight and do not acquire additional learning parameters.

\subsection{Integration of Self-Attention and Convolution}
\label{method}

The aforementioned observations naturally lead to an elegant integration of convolution and self-attention. As both modules share the same $1\!\times \!1$ convolution operations, we can only perform the projection once, and reuse these intermediate feature maps for different aggregation operations respectively. The illustration of our proposed mixed module, \ourmethod, is shown in Fig.\ref{fig:2}(c).

Specifically, \ourmethod \ also comprises two stages. At Stage I, input feature is projected by three $1\!\times \!1$ convolutions and reshaped into $N$ pieces, respectively. Thus, we obtain a rich set of intermediate features containing $3\!\times \!N$ feature maps.

At Stage II, they are used following different paradigms. For the self-attention path, we gather the intermediate features into $N$ groups, where each group contains three pieces of features, one from each $1\!\times \!1$ convolution. The corresponding three feature maps serve as queries, keys, and values, following the traditional multi-head self-attention modules (Eq.(\ref{sa3})).
For the convolution path with kernel size $k$, we adopt a light fully connected layer and generate $k^2$ feature maps.
Consequently, by shifting and aggregating the generated features (Eq.(\ref{conv2}),(\ref{conv3})), we process the input feature in a convolution manner, and gather information from a local receptive field like the traditional ones.

Finally, outputs from both paths are added together and the strengths are controlled by two learnable scalars:
\begin{equation}
\setlength{\abovedisplayskip}{1ex}
    F_{\rm out} = \alpha F_{\rm att} + \beta F_{\rm conv}.
\setlength{\belowdisplayskip}{1ex}
\end{equation}

\subsection{Improved Shift and Summation}
\label{improve}
As shown in Sec.\ref{method} and Fig.\ref{fig:2}, intermediate features in the convolution path follow the shift and summation operations as conducted in traditional convolution modules. Despite that they are theoretically lightweight, shifting tensors towards various directions practically breaks the data locality and is difficult to achieve vectorized implementation. This may greatly impair the actual efficiency of our module at the inference time.

As a remedy, we resort to applying \textbf{depthwise convolution with fixed kernels} as a replacement of the inefficient tensor shifts, as shown in Fig.\ref{fig:3} (b). Take ${\rm Shift}(f, -1, -1)$ as an example, shifted feature is computed as:
\begin{equation}
\setlength{\abovedisplayskip}{1ex}
    \label{shift_ex}
    \tilde{f}_{c,i,j} = f_{c, i-1, j-1}, \ \forall c,i,j,
\setlength{\belowdisplayskip}{1ex}
\end{equation}
where $c$ represents each channel of the input feature.

On the other hand, if we denote convolution kernel (kernel size $k=3$) as:
\begin{equation}
\setlength{\abovedisplayskip}{1ex}
    K_c = \left[
    \begin{array}{ccc}
    1 & 0 & 0 \\
    0 & 0 & 0 \\
    0 & 0 & 0 \\
    \end{array}
    \right],  \ \forall c,
\setlength{\belowdisplayskip}{1ex}
\end{equation}
the corresponding output can be formulated as:
\begin{align}
    f_{c,i,j}^{\rm (dwc)} &= \sum_{p, q \in \{0,1,2\}} K_{c, p, q}f_{c, i+p-\lfloor k/2 \rfloor, j+q-\lfloor k/2 \rfloor} \\
    &= f_{c, i-1, j-1} = \tilde{f}_{c,i,j}, \quad \forall c,i,j.
\end{align}
Therefore, with carefully designed kernel weights for specific shift directions, the convolution outputs are equivalent to the simple tensor shifts (Eq.(\ref{shift_ex})). To further incorporate with the summation of features from different directions, we concatenate all the input features and convolution kernels respectively, and formulate shift operation as a single group convolution, as depicted in Fig.\ref{fig:3} (c.I). This modification enables our module with higher computation efficiency.

On this basis, we additionally introduce several adaptations to enhance the flexibility of the module. As shown in Fig.\ref{fig:3} (c.II), we release the convolution kernel as learnable weights, with shift kernels as initialization. This improves the model capacity while maintaining the ability of original shift operations. We also use multiple groups of convolution kernels to match the output channel dimension of convolution and self-attention paths, as depicted in Fig.\ref{fig:3} (c.III).

\subsection{Computational Cost of \ourmethod}
For better comparison, we summarize the FLOPs and parameters of \ourmethod \ in Tab.\ref{flops}. The computational cost and training parameters at Stage I are the same as self-attention and lighter than traditional convolution (e.g., $3\!\times \!3$ conv). At Stage II, \ourmethod \ introduces additional computation overhead with a light fully connected layer and a group convolution described in Sec.\ref{improve}, whose computation complexity is linear with regard to channel size $C$ and comparably minor with Stage I. The practical cost in a ResNet50 model shows similar trends with theoretical analysis.

\begin{figure*}
\begin{minipage}{0.55\linewidth}
 \centerline{\includegraphics[width=1.0\linewidth]{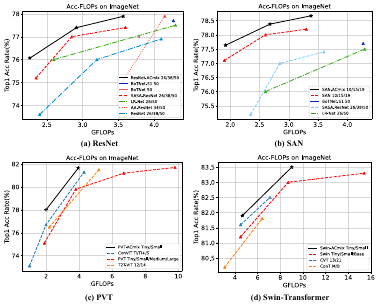}}
\end{minipage}
\hfill
\begin{minipage}{.4\linewidth}
\renewcommand\arraystretch{0.78}
\setlength{\tabcolsep}{1.0mm}{
\begin{tabular}{l|cc|l}
\toprule
\textbf{Method} & \textbf{Params} & \textbf{Flops} & \textbf{Top-1}\\
\midrule
ResNet 26  & 13.7M & 2.4G & 73.6\\
\textbf{ResNet-ACmix 26} & 10.6M & 2.3G & \textbf{76.1\,{\scriptsize (+2.5)}}\\
ResNet 38 & 19.6M & 3.2G & 76.0\\
\textbf{ResNet-ACmix 38} & 14.6M & 2.9G & \textbf{77.4\,{\scriptsize (+1.4)}}\\
ResNet 50 & 25.6M & 4.1G & 76.9\\
\textbf{ResNet-ACmix 50} & 18.6M & 3.6G & \textbf{77.8\,{\scriptsize (+0.9)}}\\
\midrule
SAN 10 & 11.8M & 1.9G & 77.1\\
\textbf{SAN-ACmix 10} & 12.1M & 1.9G & \textbf{77.6\,{\scriptsize (+0.5)}}\\
SAN 15 & 16.2M & 2.6G & 78.0\\
\textbf{SAN-ACmix 15} & 16.6M & 2.7G & \textbf{78.4\,{\scriptsize (+0.4)}}\\
SAN 19 & 20.5M & 3.3G & 78.2\\
\textbf{SAN-ACmix 19} & 21.2M & 3.4G & \textbf{78.7\,{\scriptsize (+0.5)}}\\
\midrule
PVT-T  & 13M & 1.9G & 75.1\\
\textbf{PVT-ACmix-T} & 13M & 2.0G & \textbf{78.0\,{\scriptsize (+2.9)}}\\
PVT-S  & 25M & 3.8G & 79.8\\
\textbf{PVT-ACmix-S} & 25M & 3.9G & \textbf{81.7\,{\scriptsize (+1.9)}}\\
\midrule
Swin-T & 29M & 4.5G & 81.3\\
\textbf{Swin-ACmix-T} & 30M & 4.6G & \textbf{81.9\,{\scriptsize (+0.6)}}\\
Swin-S & 50M & 8.7G & 83.0\\
\textbf{Swin-ACmix-S} & 51M & 9.0G & \textbf{83.5\,{\scriptsize (+0.5)}}\\
\bottomrule
\end{tabular}}
\end{minipage}
\vskip -0.1in
\caption{Comparisons of FLOPS and parameters against accuracy on ImageNet classification task. 
Methods in (a) adapts from ResNet-50 with tradition attentions, methods in (b) adapts from SAN with patchwise attentions, methods in (c) adapts from PVT with global attentions, and methods in (d) adapts from Swin-Transformer with window attentions.}
\label{main}
\vskip -0.2in
\end{figure*}

\subsection{Generalization to Other Attention Modes}
With the development of the self-attention mechanism, numerous researches have focused on exploring variations of the attention operator to further promote the model performance. Patchwise attention proposed by \cite{zhao2020exploring} incorporates information from all features in the local region as the attention weights to replace the original softmax operation. Window attention adopted by Swin-Transformer \cite{liu2021swin} keeps the same receptive field for tokens in the same local window to save computational cost and achieve fast inference speed. ViT and DeiT \cite{dosovitskiy2020image, touvron2021training}, on the other hand, consider global attention to retaining long-range dependencies within a single layer. These modifications are proved to be effective under specific model architectures.

Under the circumstance, it is worth noticing that our proposed \ourmethod \ is independent of self-attention formulations, and can be readily adopted on the aforementioned variants. Specifically, the attention weights can be summarized as:
\begin{align}
\setlength{\abovedisplayskip}{1ex}
    \label{patch}
    \textbf{(Patchwise)} \ &{\rm A}(q_{ij},k_{ab})\!=\!\phi([q_{ij},\![k_{ab}]_{a,b\in \mathcal{N}_k(i,j)}]), \\
    \label{window}
    \textbf{(Window)} \ \ &{\rm A}(q_{ij},k_{ab})\! =\! {\underset{a,b\in \mathcal{W}_k(i,j)}{\rm softmax}}\!\left(q_{ij}^{\rm T}k_{ab}/\sqrt{d}\right)\!, \\
    \label{global}
    \textbf{(Global)} \ \ \ &{\rm A}(q_{ij},k_{ab})\! =\! {\underset{a,b\in \mathcal{W}}{\rm softmax}}\left(q_{ij}^{\rm T}k_{ab}/\sqrt{d}\right),
\setlength{\belowdisplayskip}{1ex}
\end{align}
where $[\cdot]$ refers to feature concatenation, $\phi(\cdot)$ represents two linear projection layers with an intermediate nonlinear activation, $\mathcal{W}_k(i,j)$ is the specialized receptive field for each query token, and $\mathcal{W}$ represents the whole feature map (Please refer to the original paper for further details). Then, the computed attention weights can be applied to Eq.(\ref{sa3}) and fits into the general formulation.

\section{Experiments}
In this section, we empirically validate \ourmethod \ on ImageNet classification, semantic segmentation, and object detection tasks, and compare with state-of-the-art models. See Appendix for detailed dataset and training configurations.

\subsection{ImageNet Classification}
\noindent
\textbf{Implementation.} We practically implement \ourmethod \ on 4 baseline models, including ResNet \cite{he2016deep}, SAN \cite{zhao2020exploring}, PVT \cite{wang2021pyramid} and Swin-Transformer \cite{liu2021swin}. 
We also compare our models with competitive baselines, i.e., SASA \cite{ramachandran2019stand}, LR-Net \cite{hu2019local}, AA-ResNet \cite{bello2019attention}, BoTNet \cite{srinivas2021bottleneck}, T2T-ViT \cite{yuan2021tokens}, ConViT \cite{d2021convit}, CVT \cite{wu2021cvt}, ConT \cite{yan2021contnet} and Conformer \cite{peng2021conformer}.



\noindent
\textbf{Results.} We show the classification results in Fig.\ref{main}. For ResNet-\ourmethod \ models, our model outperforms all baselines with comparable FLOPs or parameters. For example, ResNet-\ourmethod \ 26 achieves same top-1 accuracy as SASA-ResNet 50 with $80\%$ FLOPs. With similar FLOPs, our model outperforms SASA by $0.35\% \!-\! 0.8\%$. The superiority against other baselines is even larger. For SAN-\ourmethod, PVT-\ourmethod \ and Swin-\ourmethod, our models achieve consistent improvements. As a showcase, SAN-\ourmethod \ 15 outperforms SAN 19 with $80\%$ FLOPs. PVT-\ourmethod-T shows comparable performance with PVT-Large, with only $40\%$ FLOPs. Swin-\ourmethod-S achieves higher accuracy than Swin-B with $60\%$ FLOPs.

\subsection{Downstream Tasks}
\noindent
\textbf{Semantic Segmentation}
We evaluate the effectiveness of our models on a challenging scene parsing dataset, ADE20K \cite{zhou2017scene}, and display the results on two segmentation approaches, Semantic-FPN \cite{kirillov2019panoptic} and UperNet \cite{xiao2018unified}. Backbones are pretrained on ImageNet-1K. It is shown that \ourmethod \ achieves improvements under all settings.

\noindent
\textbf{Object Detection}
We also conduct experiments on the COCO benchmark \cite{lin2014microsoft}. Tab.\ref{tab:coco} and Tab.\ref{tab:coco2} display the result of ResNet-based models and Transformer-based models with various detection heads, including RetinaNet \cite{lin2017focal}, Mask R-CNN \cite{he2017mask} and Cascade Mask R-CNN \cite{cai2018cascade}. We can observe that \ourmethod \ consistently outperform baselines with similar parameters or FLOPs. This further validate the effectiveness of \ourmethod \ when transfered to downstream tasks.


\subsection{Practical Inference Speed}
We further investigate the practical inference speed of our method under an Ascend 910 environment with MindSpore, a deep learning computing framework for mobile, edge, and cloud scenarios. We summarize the results in Tab.\ref{tab:fps}. Comparing to PVT-S, our model achieves 1.3x fps with comparable mAP. When it comes to the larger model, the superiority is more distinct. \ourmethod \ outperforms PVT-L 1.9mAP with 1.8x fps.

\begin{table}[t]
\newcommand{\tabincell}[2]{\begin{tabular}{@{}#1@{}}#2\end{tabular}}
\begin{center}
\setlength{\tabcolsep}{0.6mm}{
\renewcommand\arraystretch{0.7}
\begin{tabular}{c|c|c|cc|l}
\toprule
Method & Backbone & Schd & Params & Flops & val mIoU\\
\midrule
\multirow{4}{*}{Semantic FPN} & PVT-T & 40k & 17M & 158G & 37.1\\
 & \textbf{ACmix} & 40k & 17M & 160G & \textbf{42.7\,{\scriptsize (+5.6)}}\\
\cmidrule{2-6}
 & PVT-S & 40k & 28M & 225G & 42.4\\
 & \textbf{ACmix} & 40k & 29M & 228G & \textbf{46.4\,{\scriptsize (+4.0)}}\\
\midrule
\multirow{4}{*}{UperNet} & Swin-T & 160k & 60M & 945G & 44.5\\
 & \textbf{ACmix} & 160k & 60M & 950G & \textbf{45.3\,{\scriptsize (+0.8)}}\\
\cmidrule{2-6}
 & Swin-S & 160k & 81M & 1038G & 47.6\\
 & \textbf{ACmix} & 160k & 81M & 1043G & \textbf{48.7\,{\scriptsize (+1.1)}}\\
\bottomrule
\end{tabular}}
\end{center}
\vskip -0.2in
\caption{ADE20K segmentation with Transformer-based models. 
}
\label{tab:seg}
\vskip -0.05in
\end{table}

\begin{table}[t]
\newcommand{\tabincell}[2]{\begin{tabular}{@{}#1@{}}#2\end{tabular}}
\begin{center}
\setlength{\tabcolsep}{0.6mm}{
\renewcommand\arraystretch{0.7}
\begin{tabular}{c|c|c|c|ccc}
\toprule
Method                     & Backbone & Schd & Flops & $\mathrm{mAP}$   & $\mathrm{mAP_{50}}$ & $\mathrm{mAP_{75}}$\\
\midrule
\multirow{3}{*}{RetinaNet} & ResNet 50 & 1x & $\mathrm{250G}$ & $\mathrm{36.7}$ & $\mathrm{56.0}$  & $\mathrm{39.0}$\\
& SASA & 1x & 226G & 36.8 & 54.6 & 39.3\\
& \textbf{ACmix} & 1x & 230G & $\mathbf{38.3}$ & $\mathbf{56.2}$ & $\mathbf{40.0}$ \\
\midrule
\multirow{2}{*}{RetinaNet} & SAN 19 & 1x & 229G & 38.2 & 56.0 & 41.1\\
 & \textbf{ACmix}    & 1x  &  233G  & $\mathbf{39.1}$ & $\mathbf{58.9}$  & $\mathbf{41.6}$\\
\bottomrule
\end{tabular}}
\end{center}
\vskip -0.15in
\caption{COCO Object detection with ResNet-based models. 
}
\label{tab:coco}
\vskip -0.1in
\end{table}

\begin{table}[t]
\newcommand{\tabincell}[2]{\begin{tabular}{@{}#1@{}}#2\end{tabular}}
\begin{center}
\setlength{\tabcolsep}{0.4mm}{
\renewcommand\arraystretch{0.7}
\begin{tabular}{c|c|c|c|ccc}
\toprule
Method                     & Backbone & Schd & Flops & $\mathrm{mAP}$   & $\mathrm{mAP_{50}}$ & $\mathrm{mAP_{75}}$\\
\midrule
\multirow{2}{*}{RetinaNet} & PVT-T  & 1x & 230G & 36.7 & 56.9 & 38.9\\
 & \textbf{ACmix} & 1x & 232G & $\mathbf{40.5}$ & $\mathbf{61.2}$ & $\mathbf{42.7}$\\
\midrule
\multirow{2}{*}{RetinaNet} & PVT-T & 3x & 230G & 39.4 & 59.8 & 42.0\\
 & \textbf{ACmix} & 3x & 232G & $\mathbf{42.0}$ & $\mathbf{62.8}$ & $\mathbf{44.6}$\\
\midrule
Mask & Swin-T & 3x & 272G & 46.0 & 67.8 & 50.4\\
R-CNN & \textbf{ACmix} & 3x & 275G & $\mathbf{47.0}$ & $\mathbf{69.0}$ & $\mathbf{51.8}$\\
\midrule
Cascade & Swin-T & 3x & 750G & 50.5 & 69.3 & 54.9\\
Mask R-CNN & \textbf{ACmix} & 3x & 754G & $\mathbf{51.1}$ & $\mathbf{69.8}$ & $\mathbf{55.6}$\\
\bottomrule
\end{tabular}}
\end{center}
\vskip -0.2in
\caption{COCO Object detection with Transformer-based models. 
}
\label{tab:coco2}
\vskip -0.1in
\end{table}

\begin{table}[t]
\newcommand{\tabincell}[2]{\begin{tabular}{@{}#1@{}}#2\end{tabular}}
\begin{center}
\setlength{\tabcolsep}{2mm}{
\renewcommand\arraystretch{0.7}
\begin{tabular}{c|c|c|l}
\toprule
\bf{Method}                     & \bf{Backbone} & \bf{mAP} & \bf{FPS}\\
\midrule
\multirow{2}{*}{RetinaNet} & PVT-S & \bf{42.2} & 50.3\\
& \textbf{ACmix} & \bf{42.0} & \bf{67.7 \ {\scriptsize (x1.3)}}\\
\midrule
\multirow{2}{*}{RetinaNet} & PVT-L & 43.4 & 24.3\\
 & \textbf{ACmix} & \bf{45.3} & \bf{43.9 \ {\scriptsize (x1.8)}}\\
\bottomrule
\end{tabular}}
\end{center}
\vskip -0.2in
\caption{Practical inference speed on COCO. FPS is test on a single Ascend 910 with input image size (3, 576, 576).}
\label{tab:fps}
\vskip -0.1in
\end{table}

\begin{table}[t]
    \begin{center}
    \setlength{\tabcolsep}{1.0mm}{
    \renewcommand\arraystretch{0.7}
    \begin{tabular}{c|cc|ccc}
    \toprule
    \textbf{Method} & $\boldsymbol{\alpha}$ & $\boldsymbol{\beta}$ & \textbf{Params} & \textbf{Flops} & \textbf{top-1} \\
    \midrule
    Swin-T & 1 & - & 29M & 4.5G & 81.3\\
    \midrule
    Conv-Swin-T & - & 1 & 39M & 4.5G & 80.5\\
    \midrule
    \multirow{4}{*}{\textbf{Swin-ACmix-T}} & 1 & 1 & 30M & 4.6G & 81.5\\
    & $\alpha$ & 1 & 30M & 4.6G & 81.6\\
    & $\alpha$ & 1-$\alpha$ & 30M & 4.6G & 81.5\\
    & $\alpha$ & $\beta$ & 30M & 4.6G & \textbf{81.9}\\
    \bottomrule
    \end{tabular}}
    \end{center}
    \vskip -0.2in
    \caption{Ablation study on combining methods of two paths. The final output is computed as $F_{\rm out} = \alpha \cdot F_{\rm att} + \beta \cdot F_{\rm conv}$.}
    \label{ab-rate}
    \vskip -0.1in
\end{table}

\begin{table}[t]
    \begin{center}
    \setlength{\tabcolsep}{0.8mm}{
    \renewcommand\arraystretch{0.7}
    \begin{tabular}{c|ccc}
    \toprule
    \textbf{Shift Module} & \textbf{Flops} & \textbf{Top-1} & \textbf{FPS}\\
    \midrule
    Tensor Shifts (Fig.\ref{fig:3}a) & 4.6G & 81.4 & 313\\
    Fixed Kernel (Fig.\ref{fig:3}c.I) & 4.7G & 81.4 & 419\\
    Random Init (Fig.\ref{fig:3}c.II w/o init) & 4.7G & 81.5 & 419\\
    \midrule
    \textbf{Group Conv (Full)}& 4.7G & \textbf{81.9} & \textbf{419}\\
    \bottomrule
    \end{tabular}}
    \end{center}
    \vskip -0.2in
    \caption{Ablation study of shift modules implementations based on Swin-Transformer-T. FPS is test on a single RTX2080Ti GPU with maximum batchsize.}
    \label{ab-gc}
    \vskip -0.1in
\end{table}

\subsection{Ablation Study}
To evaluate the effectiveness of different components in \ourmethod, we conduct a series of ablation studies. 

\noindent
\textbf{Combining the output of both paths.}
We explore how different combinations of the convolution and self-attention outputs influence the model performances. We conduct experiments with multiple combination methods and summarize the results in Tab.\ref{ab-rate}. We also show the performances of models adopting only one path, Swin-T for self-attention, and Conv-Swin-T for convolution by replacing the window attention with traditional $3\!\times \!3$ convolutions. As we can observe, the combination of convolution and self-attention modules consistently outperforms models with a single path. Fixing the ratio of convolution and self-attention for all operators also leads to worse performance. Comparably, using learned parameters imposes higher flexibility for \ourmethod, and the strength for convolution and self-attention paths can be adaptively adjusted according to the position of the filter in the whole network.

\noindent
\textbf{Group Convolution Kernels.}
We also conduct ablations on the choices of group convolution kernels, as we have shown in Sec.\ref{improve} and Fig.\ref{fig:3}. We empirically show the effectiveness of each adaptation, and its influence on practical inference speed in Tab.\ref{ab-gc}. By substituting the tensor shifts with group convolutions, inference speed is greatly boosted. Also, using learnable convolution kernels and carefully-designed initialization enhance model flexibility and contribute to the final performance.

\begin{figure}
    \begin{center}
    \includegraphics[width=0.88\linewidth]{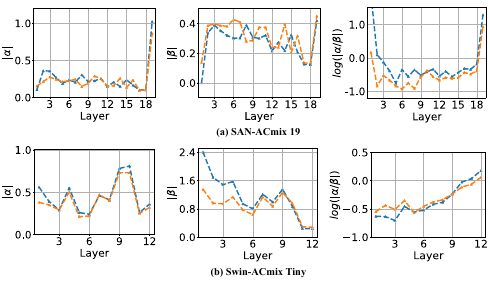}
    \vskip -0.1in
\caption{$|\alpha|$, $|\beta|$ and $log(|\alpha/\beta|)$ from different layers of SAN-\ourmethod \ and Swin-\ourmethod. Lines in the same plot correspond to parallel experiments. The final output is computed as $F_{\rm out} = \alpha \cdot F_{\rm att} + \beta \cdot F_{\rm conv}$.}
    \label{rate}
    \end{center}
    \vskip -0.3in
\end{figure}

\subsection{Bias towards Different Paths.}
It is also valuable to see that \ourmethod \ introduces two learnable scalars $\alpha, \beta$ to combine the outputs from both paths (Eq.\ref{shift_ex}). This leads to a by-product of our module, where $\alpha$ and $\beta$ practically reflect the model's bias towards convolution or self-attention at different depths. 

We conduct parallel experiments and show the learned parameters $\alpha, \beta$ from different layers of SAN-\ourmethod, and Swin-\ourmethod \ models in Fig.\ref{rate}. The left and middle plots show the changing tendency of rates for self-attention and convolution paths respectively. The variation of the rates in different experiments is relatively small, especially when layers go deeper. This observation shows a stable preference for deep models towards the different design patterns. A more distinct trend is shown in the right plot, where the ratio between two paths is explicitly presented. We can see that convolution can serve as good feature extractors at the early stages of the Transformer models. At the middle stage of the network, the model tends to leverage the mixture of both paths with an increasing bias towards convolution. At the last stage, self-attention shows superiority over convolution. This is also consistent with the design patterns in the previous works where self-attention is mostly adopted in the last stages to replace the original $3\!\times \!3$ convolution \cite{bello2019attention, srinivas2021bottleneck}, and convolutions at early stages are proved to be more effective for vision transformers \cite{xiao2021early}.

\section{Conclusion}
In this paper, we explore a close relationship between two powerful techniques, convolution and self-attention. By decomposing the operations of both modules, we show that they share the same computation overhead on projecting the input feature maps. On this basis, we take a step forward and propose a hybrid operator to integrate self-attention and convolution modules by sharing the same heavy operations. Extensive results on image classification and object detection benchmarks demonstrate the effectiveness and efficiency of the proposed operator.


\section*{Acknowledgements}
\noindent
This work is supported in part by the National Science and Technology Major Project of the Ministry of Science and Technology of China under Grants 2018AAA0100701, the National Natural Science Foundation of China under Grants 61906106 and 62022048, and Huawei Technologies Ltd.

{\small
\bibliographystyle{ieee_fullname}
\bibliography{egbib}

\begin{thebibliography}{10}\itemsep=-1pt

\bibitem{bahdanau2014neural}
Dzmitry Bahdanau, Kyunghyun Cho, and Yoshua Bengio.
\newblock Neural machine translation by jointly learning to align and
  translate.
\newblock {\em arXiv preprint arXiv:1409.0473}, 2014.

\bibitem{Beal2020TowardTO}
Josh Beal, Eric Kim, E. Tzeng, Dong~Huk Park, Andrew Zhai, and Dmitry Kislyuk.
\newblock Toward transformer-based object detection.
\newblock {\em ArXiv}, abs/2012.09958, 2020.

\bibitem{bello2019attention}
Irwan Bello, Barret Zoph, Ashish Vaswani, Jonathon Shlens, and Quoc~V Le.
\newblock Attention augmented convolutional networks.
\newblock In {\em Proceedings of the IEEE/CVF International Conference on
  Computer Vision}, pages 3286--3295, 2019.

\bibitem{NEURIPS2020_1457c0d6}
Tom Brown, Benjamin Mann, Nick Ryder, Melanie Subbiah, Jared~D Kaplan, Prafulla
  Dhariwal, Arvind Neelakantan, Pranav Shyam, Girish Sastry, Amanda Askell,
  Sandhini Agarwal, Ariel Herbert-Voss, Gretchen Krueger, Tom Henighan, Rewon
  Child, Aditya Ramesh, Daniel Ziegler, Jeffrey Wu, Clemens Winter, Chris
  Hesse, Mark Chen, Eric Sigler, Mateusz Litwin, Scott Gray, Benjamin Chess,
  Jack Clark, Christopher Berner, Sam McCandlish, Alec Radford, Ilya Sutskever,
  and Dario Amodei.
\newblock Language models are few-shot learners.
\newblock In H. Larochelle, M. Ranzato, R. Hadsell, M.~F. Balcan, and H. Lin,
  editors, {\em Advances in Neural Information Processing Systems}, volume~33,
  pages 1877--1901. Curran Associates, Inc., 2020.

\bibitem{cai2018cascade}
Zhaowei Cai and Nuno Vasconcelos.
\newblock Cascade r-cnn: Delving into high quality object detection.
\newblock In {\em Proceedings of the IEEE conference on computer vision and
  pattern recognition}, pages 6154--6162, 2018.

\bibitem{Cao2019GCNetNN}
Yue Cao, J. Xu, Stephen Lin, Fangyun Wei, and H. Hu.
\newblock Gcnet: Non-local networks meet squeeze-excitation networks and
  beyond.
\newblock {\em 2019 IEEE/CVF International Conference on Computer Vision
  Workshop (ICCVW)}, pages 1971--1980, 2019.

\bibitem{carion2020end}
Nicolas Carion, Francisco Massa, Gabriel Synnaeve, Nicolas Usunier, Alexander
  Kirillov, and Sergey Zagoruyko.
\newblock End-to-end object detection with transformers.
\newblock In {\em European Conference on Computer Vision}, pages 213--229.
  Springer, 2020.

\bibitem{Chen2020PreTrainedIP}
Hanting Chen, Yunhe Wang, Tianyu Guo, Chang Xu, Yiping Deng, Zhenhua Liu, Siwei
  Ma, Chunjing Xu, Chao Xu, and Wen Gao.
\newblock Pre-trained image processing transformer.
\newblock In {\em Proceedings of the IEEE/CVF Conference on Computer Vision and
  Pattern Recognition (CVPR)}, pages 12299--12310, June 2021.

\bibitem{chen2017deeplab}
Liang-Chieh Chen, George Papandreou, Iasonas Kokkinos, Kevin Murphy, and Alan~L
  Yuille.
\newblock Deeplab: Semantic image segmentation with deep convolutional nets,
  atrous convolution, and fully connected crfs.
\newblock {\em IEEE transactions on pattern analysis and machine intelligence},
  40(4):834--848, 2017.

\bibitem{child2019generating}
Rewon Child, Scott Gray, Alec Radford, and Ilya Sutskever.
\newblock Generating long sequences with sparse transformers.
\newblock {\em arXiv preprint arXiv:1904.10509}, 2019.

\bibitem{cordonnier2019relationship}
Jean-Baptiste Cordonnier, Andreas Loukas, and Martin Jaggi.
\newblock On the relationship between self-attention and convolutional layers.
\newblock In {\em International Conference on Learning Representations}, 2020.

\bibitem{d2021convit}
St{\'e}phane d'Ascoli, Hugo Touvron, Matthew Leavitt, Ari Morcos, Giulio
  Biroli, and Levent Sagun.
\newblock Convit: Improving vision transformers with soft convolutional
  inductive biases.
\newblock {\em arXiv preprint arXiv:2103.10697}, 2021.

\bibitem{deng2009imagenet}
Jia Deng, Wei Dong, Richard Socher, Li-Jia Li, Kai Li, and Li Fei-Fei.
\newblock Imagenet: A large-scale hierarchical image database.
\newblock In {\em 2009 IEEE conference on computer vision and pattern
  recognition}, pages 248--255. Ieee, 2009.

\bibitem{devlin2018bert}
Jacob Devlin, Ming{-}Wei Chang, Kenton Lee, and Kristina Toutanova.
\newblock {BERT:} pre-training of deep bidirectional transformers for language
  understanding.
\newblock In Jill Burstein, Christy Doran, and Thamar Solorio, editors, {\em
  Proceedings of the 2019 Conference of the North American Chapter of the
  Association for Computational Linguistics: Human Language Technologies,
  {NAACL-HLT} 2019, Minneapolis, MN, USA, June 2-7, 2019, Volume 1 (Long and
  Short Papers)}, pages 4171--4186. Association for Computational Linguistics,
  2019.

\bibitem{dong2021cswin}
Xiaoyi Dong, Jianmin Bao, Dongdong Chen, Weiming Zhang, Nenghai Yu, Lu Yuan,
  Dong Chen, and Baining Guo.
\newblock Cswin transformer: A general vision transformer backbone with
  cross-shaped windows.
\newblock {\em arXiv preprint arXiv:2107.00652}, 2021.

\bibitem{dosovitskiy2020image}
Alexey Dosovitskiy, Lucas Beyer, Alexander Kolesnikov, Dirk Weissenborn,
  Xiaohua Zhai, Thomas Unterthiner, Mostafa Dehghani, Matthias Minderer, Georg
  Heigold, Sylvain Gelly, Jakob Uszkoreit, and Neil Houlsby.
\newblock An image is worth 16x16 words: Transformers for image recognition at
  scale.
\newblock In {\em International Conference on Learning Representations}, 2021.

\bibitem{gao2021container}
Peng Gao, Jiasen Lu, Hongsheng Li, Roozbeh Mottaghi, and Aniruddha Kembhavi.
\newblock Container: Context aggregation network.
\newblock {\em arXiv preprint arXiv:2106.01401}, 2021.

\bibitem{Guo2020PCTPC}
Menghao Guo, Jun-Xiong Cai, Zheng-Ning Liu, Tai-Jiang Mu, R. Martin, and S. Hu.
\newblock Pct: Point cloud transformer.
\newblock {\em ArXiv}, abs/2012.09688, 2020.

\bibitem{he2017mask}
Kaiming He, Georgia Gkioxari, Piotr Doll{\'a}r, and Ross Girshick.
\newblock Mask r-cnn.
\newblock In {\em Proceedings of the IEEE international conference on computer
  vision}, pages 2961--2969, 2017.

\bibitem{he2016deep}
Kaiming He, Xiangyu Zhang, Shaoqing Ren, and Jian Sun.
\newblock Deep residual learning for image recognition.
\newblock In {\em Proceedings of the IEEE conference on computer vision and
  pattern recognition}, pages 770--778, 2016.

\bibitem{hu2019local}
Han Hu, Zheng Zhang, Zhenda Xie, and Stephen Lin.
\newblock Local relation networks for image recognition.
\newblock In {\em Proceedings of the IEEE/CVF International Conference on
  Computer Vision}, pages 3464--3473, 2019.

\bibitem{hu2018gather}
Jie Hu, Li Shen, Samuel Albanie, Gang Sun, and Andrea Vedaldi.
\newblock Gather-excite: Exploiting feature context in convolutional neural
  networks.
\newblock 2018.

\bibitem{hu2018squeeze}
Jie Hu, Li Shen, and Gang Sun.
\newblock Squeeze-and-excitation networks.
\newblock In {\em Proceedings of the IEEE conference on computer vision and
  pattern recognition}, pages 7132--7141, 2018.

\bibitem{huang2019convolutional}
Gao Huang, Zhuang Liu, Geoff Pleiss, Laurens Van Der~Maaten, and Kilian
  Weinberger.
\newblock Convolutional networks with dense connectivity.
\newblock {\em IEEE transactions on pattern analysis and machine intelligence},
  2019.

\bibitem{huang2017densely}
Gao Huang, Zhuang Liu, Laurens Van Der~Maaten, and Kilian~Q Weinberger.
\newblock Densely connected convolutional networks.
\newblock In {\em Proceedings of the IEEE conference on computer vision and
  pattern recognition}, pages 4700--4708, 2017.

\bibitem{kirillov2019panoptic}
Alexander Kirillov, Ross Girshick, Kaiming He, and Piotr Doll{\'a}r.
\newblock Panoptic feature pyramid networks.
\newblock In {\em Proceedings of the IEEE/CVF Conference on Computer Vision and
  Pattern Recognition}, pages 6399--6408, 2019.

\bibitem{Krizhevsky2012ImageNetCW}
A. Krizhevsky, Ilya Sutskever, and Geoffrey~E. Hinton.
\newblock Imagenet classification with deep convolutional neural networks.
\newblock {\em Communications of the ACM}, 60:84 -- 90, 2012.

\bibitem{LeCun1989BackpropagationAT}
Y. LeCun, B. Boser, J. Denker, D. Henderson, R. Howard, W. Hubbard, and L.
  Jackel.
\newblock Backpropagation applied to handwritten zip code recognition.
\newblock {\em Neural Computation}, 1:541--551, 1989.

\bibitem{Li2019SelectiveKN}
Xiang Li, Wenhai Wang, Xiaolin Hu, and Jian Yang.
\newblock Selective kernel networks.
\newblock {\em 2019 IEEE/CVF Conference on Computer Vision and Pattern
  Recognition (CVPR)}, pages 510--519, 2019.

\bibitem{lin2017focal}
Tsung-Yi Lin, Priya Goyal, Ross Girshick, Kaiming He, and Piotr Doll{\'a}r.
\newblock Focal loss for dense object detection.
\newblock In {\em Proceedings of the IEEE international conference on computer
  vision}, pages 2980--2988, 2017.

\bibitem{lin2014microsoft}
Tsung-Yi Lin, Michael Maire, Serge Belongie, James Hays, Pietro Perona, Deva
  Ramanan, Piotr Doll{\'a}r, and C~Lawrence Zitnick.
\newblock Microsoft {COCO}: Common objects in context.
\newblock In {\em ECCV}, 2014.

\bibitem{liu2021swin}
Ze Liu, Yutong Lin, Yue Cao, Han Hu, Yixuan Wei, Zheng Zhang, Stephen Lin, and
  Baining Guo.
\newblock Swin transformer: Hierarchical vision transformer using shifted
  windows.
\newblock {\em arXiv preprint arXiv:2103.14030}, 2021.

\bibitem{pan20203d}
Xuran Pan, Zhuofan Xia, Shiji Song, Li~Erran Li, and Gao Huang.
\newblock 3d object detection with pointformer.
\newblock In {\em Proceedings of the IEEE/CVF Conference on Computer Vision and
  Pattern Recognition (CVPR)}, pages 7463--7472, June 2021.

\bibitem{park2018bam}
Jongchan Park, Sanghyun Woo, Joon-Young Lee, and In~So Kweon.
\newblock Bam: Bottleneck attention module.
\newblock {\em arXiv preprint arXiv:1807.06514}, 2018.

\bibitem{parmar2018image}
Niki Parmar, Ashish Vaswani, Jakob Uszkoreit, Lukasz Kaiser, Noam Shazeer,
  Alexander Ku, and Dustin Tran.
\newblock Image transformer.
\newblock In {\em International Conference on Machine Learning}, pages
  4055--4064. PMLR, 2018.

\bibitem{peng2021conformer}
Zhiliang Peng, Wei Huang, Shanzhi Gu, Lingxi Xie, Yaowei Wang, Jianbin Jiao,
  and Qixiang Ye.
\newblock Conformer: Local features coupling global representations for visual
  recognition.
\newblock {\em arXiv preprint arXiv:2105.03889}, 2021.

\bibitem{radford2018improving}
Alec Radford, Karthik Narasimhan, Tim Salimans, and Ilya Sutskever.
\newblock Improving language understanding by generative pre-training.
\newblock 2018.

\bibitem{ramachandran2019stand}
Prajit Ramachandran, Niki Parmar, Ashish Vaswani, Irwan Bello, Anselm Levskaya,
  and Jon Shlens.
\newblock Stand-alone self-attention in vision models.
\newblock In H. Wallach, H. Larochelle, A. Beygelzimer, F. d\textquotesingle
  Alch\'{e}-Buc, E. Fox, and R. Garnett, editors, {\em Advances in Neural
  Information Processing Systems}, volume~32. Curran Associates, Inc., 2019.

\bibitem{ren2015faster}
Shaoqing Ren, Kaiming He, Ross Girshick, and Jian Sun.
\newblock Faster r-cnn: Towards real-time object detection with region proposal
  networks.
\newblock {\em Advances in neural information processing systems}, 28:91--99,
  2015.

\bibitem{Simonyan2015VeryDC}
K. Simonyan and Andrew Zisserman.
\newblock Very deep convolutional networks for large-scale image recognition.
\newblock {\em CoRR}, abs/1409.1556, 2015.

\bibitem{srinivas2021bottleneck}
Aravind Srinivas, Tsung-Yi Lin, Niki Parmar, Jonathon Shlens, Pieter Abbeel,
  and Ashish Vaswani.
\newblock Bottleneck transformers for visual recognition.
\newblock In {\em Proceedings of the IEEE/CVF Conference on Computer Vision and
  Pattern Recognition}, pages 16519--16529, 2021.

\bibitem{touvron2021training}
Hugo Touvron, Matthieu Cord, Matthijs Douze, Francisco Massa, Alexandre
  Sablayrolles, and Herv{\'e} J{\'e}gou.
\newblock Training data-efficient image transformers \& distillation through
  attention.
\newblock In {\em International Conference on Machine Learning}, pages
  10347--10357. PMLR, 2021.

\bibitem{vaswani2017attention}
Ashish Vaswani, Noam Shazeer, Niki Parmar, Jakob Uszkoreit, Llion Jones,
  Aidan~N Gomez, {\L}ukasz Kaiser, and Illia Polosukhin.
\newblock Attention is all you need.
\newblock In {\em Advances in neural information processing systems}, pages
  5998--6008, 2017.

\bibitem{wang2021pyramid}
Wenhai Wang, Enze Xie, Xiang Li, Deng-Ping Fan, Kaitao Song, Ding Liang, Tong
  Lu, Ping Luo, and Ling Shao.
\newblock Pyramid vision transformer: A versatile backbone for dense prediction
  without convolutions.
\newblock {\em arXiv preprint arXiv:2102.12122}, 2021.

\bibitem{wang2018non}
Xiaolong Wang, Ross Girshick, Abhinav Gupta, and Kaiming He.
\newblock Non-local neural networks.
\newblock In {\em Proceedings of the IEEE conference on computer vision and
  pattern recognition}, pages 7794--7803, 2018.

\bibitem{Wang2020EndtoEndVI}
Yuqing Wang, Zhaoliang Xu, Xinlong Wang, Chunhua Shen, Baoshan Cheng, Hao Shen,
  and Huaxia Xia.
\newblock End-to-end video instance segmentation with transformers.
\newblock In {\em Proceedings of the IEEE/CVF Conference on Computer Vision and
  Pattern Recognition}, pages 8741--8750, 2021.

\bibitem{woo2018cbam}
Sanghyun Woo, Jongchan Park, Joon-Young Lee, and In~So Kweon.
\newblock Cbam: Convolutional block attention module.
\newblock In {\em Proceedings of the European conference on computer vision
  (ECCV)}, pages 3--19, 2018.

\bibitem{wu2021cvt}
Haiping Wu, Bin Xiao, Noel Codella, Mengchen Liu, Xiyang Dai, Lu Yuan, and Lei
  Zhang.
\newblock Cvt: Introducing convolutions to vision transformers.
\newblock {\em arXiv preprint arXiv:2103.15808}, 2021.

\bibitem{xiao2018unified}
Tete Xiao, Yingcheng Liu, Bolei Zhou, Yuning Jiang, and Jian Sun.
\newblock Unified perceptual parsing for scene understanding.
\newblock In {\em Proceedings of the European Conference on Computer Vision
  (ECCV)}, pages 418--434, 2018.

\bibitem{xiao2021early}
Tete Xiao, Mannat Singh, Eric Mintun, Trevor Darrell, Piotr Doll{\'a}r, and
  Ross Girshick.
\newblock Early convolutions help transformers see better.
\newblock {\em arXiv preprint arXiv:2106.14881}, 2021.

\bibitem{yan2021contnet}
Haotian Yan, Zhe Li, Weijian Li, Changhu Wang, Ming Wu, and Chuang Zhang.
\newblock Contnet: Why not use convolution and transformer at the same time?
\newblock {\em arXiv preprint arXiv:2104.13497}, 2021.

\bibitem{yuan2021tokens}
Li Yuan, Yunpeng Chen, Tao Wang, Weihao Yu, Yujun Shi, Zihang Jiang, Francis~EH
  Tay, Jiashi Feng, and Shuicheng Yan.
\newblock Tokens-to-token vit: Training vision transformers from scratch on
  imagenet.
\newblock {\em arXiv preprint arXiv:2101.11986}, 2021.

\bibitem{Zhang2020FeaturePT}
Dong Zhang, Hanwang Zhang, J. Tang, Meng Wang, Xiansheng Hua, and Qianru Sun.
\newblock Feature pyramid transformer.
\newblock {\em ArXiv}, abs/2007.09451, 2020.

\bibitem{zhao2020exploring}
Hengshuang Zhao, Jiaya Jia, and Vladlen Koltun.
\newblock Exploring self-attention for image recognition.
\newblock In {\em Proceedings of the IEEE/CVF Conference on Computer Vision and
  Pattern Recognition}, pages 10076--10085, 2020.

\bibitem{zheng2020rethinking}
Sixiao Zheng, Jiachen Lu, Hengshuang Zhao, Xiatian Zhu, Zekun Luo, Yabiao Wang,
  Yanwei Fu, Jianfeng Feng, Tao Xiang, Philip~HS Torr, et~al.
\newblock Rethinking semantic segmentation from a sequence-to-sequence
  perspective with transformers.
\newblock In {\em Proceedings of the IEEE/CVF Conference on Computer Vision and
  Pattern Recognition}, pages 6881--6890, 2021.

\bibitem{zhou2017scene}
Bolei Zhou, Hang Zhao, Xavier Puig, Sanja Fidler, Adela Barriuso, and Antonio
  Torralba.
\newblock Scene parsing through ade20k dataset.
\newblock In {\em Proceedings of the IEEE conference on computer vision and
  pattern recognition}, pages 633--641, 2017.

\bibitem{Zhu2020DeformableDD}
Xizhou Zhu, Weijie Su, Lewei Lu, Bin Li, Xiaogang Wang, and Jifeng Dai.
\newblock Deformable detr: Deformable transformers for end-to-end object
  detection.
\newblock In {\em International Conference on Learning Representations}, 2021.

\end{thebibliography}
}
\clearpage
\section*{Appendix}

\section*{A. Model Architectures}
We summarize the architectures of ResNet 26/38/50 \cite{he2016deep}, SAN 10/15/19 \cite{zhao2020exploring}, PVT-T/S \cite{wang2021pyramid}, Swin-T/S \cite{liu2021swin}, and their respective \ourmethod \ version in Tab \ref{Tab01}$\sim$\ref{Tab04}. For fair comparison, we only substitute the original $3\!\times \!3$ convolution or self-attention module with our proposed operator in the modified models.

\section*{B. Dataset and Training Setup}
\noindent
\textbf{ImageNet.} ImageNet 2012 \cite{deng2009imagenet} comprises 1.28 million training images and 50,000 validation images from 1000 different classes. For ResNet-based models, we follow the training schedule in \cite{zhao2020exploring} and train all the models for 100 epochs. We use SGD with batchsize 256 on 8 GPUs. Cosine learning rate is adopted with the base learning rate set to 0.1. We apply standard data augmentation, including random cropping, random horizontal flipping and normalization. We use label smoothing with coefficient 0.1. For experiments on Transformer-based models, including PVT and Swin-Transformer, we follow training configurations in the original paper.

\noindent
\textbf{COCO.} COCO dataset \cite{lin2014microsoft} is a standard object detection benchmark and we use a subset of 80k samples as training set and 35k for validation. For ResNet and SAN models, we train the network by SGD and 8 GPU are used with a batchsize of 16. For PVT and Swin-Transformer models, we train the network by adamw. Backbone networks are respectively pretrained on ImageNet dataset following the same training configurations in the original paper. We follow the "1x" learning schedule to train the whole network for 12 epochs and divide the learning rate by 10 at the 8th and 11th epoch respectively. For several transformer-based models, we follow the configurations in the original paper, and additionally experiment "3x" schedule with 36 epochs.
We apply standard data augmentation, that is resize, random flip and normalize. Learning rate is set at 0.01 and linear warmup is used in the first 500 iterations. We follow the "1x" learning schedule training the whole network for 12 epochs and divide the learning rate by 10 at the 8th and 11th epoch respectively. For several transformer-based models, we follow the configurations in the original paper, and test with "3x" schedule. All mAP results in the main paper are tested with input image size (3, 1333, 800).

\noindent
\textbf{ADE20K.} ADE20K \cite{zhou2017scene} is a widely-used semantic segmentation dataset, containing 150 categories. ADE20K has 25K images, with 20K for training, 2K for validation, and another 3K for testing. For two baseline models, PVT and Swin-Transformer, we follow the training configurations in their original paper respectively. For PVT, we implement the backbone models on the basis of Semantic FPN \cite{kirillov2019panoptic}. We optimize the models using AdamW with an initial learning rate of 1e-4 for 80k iterations. For Swin-Transformer, we implement the backbone models on the basis of UperNet \cite{xiao2018unified}. We use the AdamW optimizer with an initial learning rate of 6e-5 and a linear warmup of 1,500 iterations. Models are trained for a total of 160K iterations. We randomly resize and crop the image to 512 × 512 for training, and rescale to have a shorter side of 512 pixels during testing.

\section*{C. Hyper-parameters}
For ResNet-\ourmethod \ models, we set $N\!=\!4$ for all the experiments. 

For SAN-\ourmethod \ models, the channel dimension for queries, keys and values are different in the original model \cite{zhao2020exploring}. Given input features with channel dimension $C$, queries and keys are projected to features with $C/4$ channels, while values are projected to features with $C$ channels. Therefore, when implementing our \ourmethod \ operator, we divide values into 4 groups, where the divided groups have the same channel dimension $C/4$. The following self-attention and convolution operations follow the same \textit{patchwise} attention in \cite{zhao2020exploring} and the same designing pipeline as we stated in Sec.4, respectively.

For PVT-\ourmethod \ and Swin-\ourmethod \ models, we follow the configurations in the original model \cite{liu2021swin}.

$k_a\!=\!7$ and $k_c\!=\!3$ is set for all experiments, unless stated otherwise.


\section*{D. Positional Encoding}
Positional encoding is widely adopted in self-attention modules, while not used in SAN and PVT models. Therefore, we follow this setting and only adopt positional encoding in the ResNet-\ourmethod \ models and Swin-\ourmethod. Specifically, the popular relative positional encoding \cite{ramachandran2019stand} is adopted when computing the attention weights:
\begin{equation}
\label{pairwise}
    {\rm A}(q_{ij},k_{ab})={\underset{\mathcal{N}(i,j)}{\rm softmax}}\left((q_{ij}^{\rm T}k_{ab} + B_{ij,ab})/\sqrt{d}\right),
\end{equation}
where $q, k, B$ represent queries, keys and relative positional encodings respectively. We didn't include positional encoding in the analysis for computation complexity in the Tab.1 of the main paper, as the \textit{patchwise} attention proposed in \cite{zhao2020exploring} demonstrate the effectiveness of self-attention modules without adopting it. Nevertheless, the computation cost for positional encoding is also linear with respect to the channel dimension $C$, which is also comparably minor to the feature projection operations. Therefore, considering the positional encoding doesn't affect our main statement.

\section*{E. Practical Costs for Other Models}
We also summarize the practical FLOPs and Parameters for convolution, self-attention and \ourmethod \  based on various models introduced in the Experiment section. The numbers are shown in Tab.\ref{flops_app}. Similar to ResNet 50, more than $60\%$ of the computation are performed at Stage I of the self-attention module in SAN and Swin models. Meanwhile, it also demonstrates that \ourmethod \ only introduces minimum computational cost to integrate both convolution and self-attention modules based on various model structures.

\begin{table}[h]
    \begin{center}
    \setlength{\tabcolsep}{0.6mm}{
    \renewcommand\arraystretch{1.2}
    \begin{tabular}{cc|c|c|c}
    \toprule
     & &\textbf{ResNet 50} & \textbf{SAN 19} & \textbf{Swin-T}\\
    \textbf{Module} & \textbf{Stage} & \textbf{GFLOPs} & \textbf{GFLOPs} & \textbf{GFLOPs}\\
    \midrule
    \multirow{2}{*}{Convolution} & I & $1.85\ (99\%)$ & - & - \\
    & II & $0.01\ (1\%)$ & - & - \\
    \midrule
    \multirow{2}{*}{Self-Attention} & I & $0.96\ (83\%)$ & $1.29\ (64\%)$ & $1.04\ (68\%)$ \\
    & II & $0.19\ (17\%)$ & $0.72\ (36\%)$ & $0.49\ (32\%)$\\
    \midrule
    \multirow{2}{*}{\textbf{\ourmethod}} & I & $0.96\ (73\%)$ & $1.29\ (60\%)$ & $1.04\ (62\%)$\\
    & II & $0.35\ (27\%)$ & $0.89 (40\%)$ & $0.64\ (38\%)$\\
    \bottomrule
    \end{tabular}}
    \end{center}
    \caption{Practical FLOPs and Parameters for different modules based on various models. Numbers within the brackets are their fractions of the whole module. SAN 19 and Swin-T models are designed with all self-attention modules, thus not applicable for the traditional convolution module.}
    \label{flops_app}
\end{table}

\begin{table*}[t]
    \centering
    \setlength{\tabcolsep}{0.5mm}{
    \renewcommand\arraystretch{1.15}
    \begin{tabular}{c|c|c|c|c}
    \thickhline
    stage & output & ResNet-26 (\textbf{\ourmethod}) & ResNet-38 (\textbf{\ourmethod}) & ResNet-50 (\textbf{\ourmethod}) \\
    \hline
    res1 & $112\times 112$ & $7\times 7$ conv, 64, stride 2 & $7\times 7$ conv, 64, stride 2 & $7\times 7$ conv, 64, stride 2\\
    \hline
    \multirow{4}*{res2} & \multirow{4}*{$56\times 56$} & $3\times 3$ max pool, stride 2 & $3\times 3$ max pool, stride 2 & $3\times 3$ max pool, stride 2\\
    \cline{3-5}
    && $\left[ \begin{array}{c} 1\times 1 {\rm \ conv}, 64\\3
    \times 3 \ {\rm conv \ (\textbf{\ourmethod})}, 64\\1\times 1 {\rm \ conv}, 256\end{array} \right ] \!\times \!1$ & $\left[ \begin{array}{c} 1\times 1 {\rm \ conv}, 64\\3
    \times 3 \ {\rm conv \ (\textbf{\ourmethod})}, 64\\1\times 1 {\rm \ conv}, 256\end{array} \right ] \!\times \!2$ & $\left[ \begin{array}{c} 1\times 1 {\rm \ conv}, 64\\3
    \times 3 \ {\rm conv \ (\textbf{\ourmethod})}, 64\\1\times 1 {\rm \ conv}, 256\end{array} \right ] \!\times \!3$\\
    \hline
    res3 & $28 \times 28$ & $\left[ \begin{array}{c} 1\times 1 {\rm \ conv}, 128\\3
    \times 3 \ {\rm conv \ (\textbf{\ourmethod})}, 128\\1\times 1 {\rm \ conv}, 512\end{array} \right ] \!\times \!2$ & $\left[ \begin{array}{c} 1\times 1 {\rm \ conv}, 128\\3
    \times 3 \ {\rm conv \ (\textbf{\ourmethod})}, 128\\1\times 1 {\rm \ conv}, 512\end{array} \right ] \!\times \!3$ & $\left[ \begin{array}{c} 1\times 1 {\rm \ conv}, 128\\3
    \times 3 \ {\rm conv \ (\textbf{\ourmethod})}, 128\\1\times 1 {\rm \ conv}, 512\end{array} \right ] \!\times \!4$\ \\
    \hline
    res4 & $14 \times 14$ & $\left[ \begin{array}{c} 1\times 1 {\rm \ conv}, 256\\3
    \times 3 \ {\rm conv \ (\textbf{\ourmethod})}, 256\\1\times 1 {\rm \ conv}, 1024\end{array} \right ] \!\times \!4$ & $\left[ \begin{array}{c} 1\times 1 {\rm \ conv}, 256\\3
    \times 3 \ {\rm conv \ (\textbf{\ourmethod})}, 256\\1\times 1 {\rm \ conv}, 1024\end{array} \right ] \!\times \!5$ & $\left[ \begin{array}{c} 1\times 1 {\rm \ conv}, 256\\3
    \times 3 \ {\rm conv \ (\textbf{\ourmethod})}, 256\\1\times 1 {\rm \ conv}, 1024\end{array} \right ] \!\times \!6$\ \\
    \hline
    res5 & $7 \times 7$ & $\left[ \begin{array}{c} 1\times 1 {\rm \ conv}, 512\\3
    \times 3 \ {\rm conv \ (\textbf{\ourmethod})}, 512\\1\times 1 {\rm \ conv}, 2048\end{array} \right ] \!\times \!1$ & $\left[ \begin{array}{c} 1\times 1 {\rm \ conv}, 512\\3
    \times 3 \ {\rm conv \ (\textbf{\ourmethod})}, 512\\1\times 1 {\rm \ conv}, 2048\end{array} \right ] \!\times \!2$ & $\left[ \begin{array}{c} 1\times 1 {\rm \ conv}, 512\\3
    \times 3 \ {\rm conv \ (\textbf{\ourmethod})}, 512\\1\times 1 {\rm \ conv}, 2048\end{array} \right ] \!\times \!3$\ \\
    \hline
    \multirow{2}*{} & \multirow{2}*{$1\times 1$} & global average pool & global average pool & global average pool\\
    & & 1000-d fc, softmax & 1000-d fc, softmax & 1000-d fc, softmax\\
    \thickhline
    \end{tabular}}
    \vspace{2pt}
    \caption{Architectures of ResNet-based models with and without \ourmethod \ modules.}
    \label{Tab01}
\end{table*}

\begin{table*}[t]
    \centering
    \setlength{\tabcolsep}{0.5mm}{
    \renewcommand\arraystretch{1.15}
    \begin{tabular}{c|c|c|c|c}
    \thickhline
    layers & output & SAN-10 (\textbf{\ourmethod}) & SAN-15 (\textbf{\ourmethod}) & SAN-19 (\textbf{\ourmethod}) \\
    \hline
    Input & $224\times 224$ & $1\times 1$ conv, 64 & $1\times 1$ conv, 64 & $1\times 1$ conv, 64\\
    \hline
    \multirow{2}*{Transition} & \multirow{2}*{$112\times 112$} & $2\times 2$ max pool, stride 2 & $2\times 2$ max pool, stride 2 & $2\times 2$ max pool, stride 2\\
    && $1 \times 1$ conv, 64 & $1 \times 1$ conv, 64 & $1 \times 1$ conv, 64\\
    \hline
    Block & $112 \times 112$ & $\left[ \begin{array}{c} 3\times 3 {\rm \ sa \ (\textbf{\ourmethod})}, 16\\1\times 1 {\rm \ conv}, 64\end{array} \right ] \times 2$ & $\left[ \begin{array}{c} 3\times 3 {\rm \ sa \ (\textbf{\ourmethod})}, 16\\1\times 1 {\rm \ conv}, 64\end{array} \right ] \times 3$ & $\left[ \begin{array}{c} 3\times 3 {\rm \ sa \ (\textbf{\ourmethod})}, 16\\1\times 1 {\rm \ conv}, 64\end{array} \right ] \times 3$\ \\
    \hline
    \multirow{2}*{Transition} & \multirow{2}*{$56\times 56$} & $2\times 2$ max pool, stride 2 & $2\times 2$ max pool, stride 2 & $2\times 2$ max pool, stride 2\\
    && $1 \times 1$ conv, 256 & $1 \times 1$ conv, 256 & $1 \times 1$ conv, 256\\
    \hline
    Block & $56 \times 56$ & $\left[ \begin{array}{c} 7\times 7 {\rm \ sa \ (\textbf{\ourmethod})}, 64\\1\times 1 {\rm \ conv}, 256\end{array} \right ] \times 1$ & $\left[ \begin{array}{c} 7\times 7 {\rm \ sa \ (\textbf{\ourmethod})}, 64\\1\times 1 {\rm \ conv}, 256\end{array} \right ] \times 2$ & $\left[ \begin{array}{c} 7\times 7 {\rm \ sa \ (\textbf{\ourmethod})}, 64\\1\times 1 {\rm \ conv}, 256\end{array} \right ] \times 3$\ \\
    \hline
    \multirow{2}*{Transition} & \multirow{2}*{$28\times 28$} & $2\times 2$ max pool, stride 2 & $2\times 2$ max pool, stride 2 & $2\times 2$ max pool, stride 2\\
    && $1 \times 1$ conv, 512 & $1 \times 1$ conv, 512 & $1 \times 1$ conv, 512\\
    \hline
    Block & $28 \times 28$ & $\left[ \begin{array}{c} 7\times 7 {\rm \ sa \ (\textbf{\ourmethod})}, 128\\1\times 1 {\rm \ conv}, 512\end{array} \right ] \times 2$ & $\left[ \begin{array}{c} 7\times 7 {\rm \ sa \ (\textbf{\ourmethod})}, 128\\1\times 1 {\rm \ conv}, 512\end{array} \right ] \times 3$ & $\left[ \begin{array}{c} 7\times 7 {\rm \ sa \ (\textbf{\ourmethod})}, 128\\1\times 1 {\rm \ conv}, 512\end{array} \right ] \times 4$ \\
    \hline
    \multirow{2}*{Transition} & \multirow{2}*{$14\times 14$} & $2\times 2$ max pool, stride 2 & $2\times 2$ max pool, stride 2 & $2\times 2$ max pool, stride 2\\
    && $1 \times 1$ conv, 1024 & $1 \times 1$ conv, 1024 & $1 \times 1$ conv, 1024\\
    \hline
    Block & $14 \times 14$ & $\left[ \begin{array}{c} 7\times 7 {\rm \ sa \ (\textbf{\ourmethod})}, 256\\1\times 1 {\rm \ conv}, 1024\end{array} \right ] \times 4$ & $\left[ \begin{array}{c} 7\times 7 {\rm \ sa \ (\textbf{\ourmethod})}, 256\\1\times 1 {\rm \ conv}, 1024\end{array} \right ] \times 5$ & $\left[ \begin{array}{c} 7\times 7 {\rm \ sa \ (\textbf{\ourmethod})}, 256\\1\times 1 {\rm \ conv}, 1024\end{array} \right ] \times 6$ \\
    \hline
    \multirow{2}*{Transition} & \multirow{2}*{$7\times 7$} & $2\times 2$ max pool, stride 2 & $2\times 2$ max pool, stride 2 & $2\times 2$ max pool, stride 2\\
    && $1 \times 1$ conv, 2048 & $1 \times 1$ conv, 2048 & $1 \times 1$ conv, 2048\\
    \hline
    Block & $7 \times 7$ & $\left[ \begin{array}{c} 7\times 7 {\rm \ sa \ (\textbf{\ourmethod})}, 512\\1\times 1 {\rm \ conv}, 2048\end{array} \right ] \times 1$ & $\left[ \begin{array}{c} 7\times 7 {\rm \ sa \ (\textbf{\ourmethod})}, 512\\1\times 1 {\rm \ conv}, 2048\end{array} \right ] \times 2$ & $\left[ \begin{array}{c} 7\times 7 {\rm \ sa \ (\textbf{\ourmethod})}, 512\\1\times 1 {\rm \ conv}, 2048\end{array} \right ] \times 3$ \\
    \hline
    \multirow{2}*{Classification} & \multirow{2}*{$1\times 1$} & global average pool & global average pool & global average pool\\
    & & 1000-d fc, softmax & 1000-d fc, softmax & 1000-d fc, softmax\\
    \thickhline
    \end{tabular}}
    \vspace{2pt}
    \caption{Architectures of SAN-based models with and without \ourmethod \ modules.}
    \label{Tab02}
\end{table*}

\begin{table*}[t]
    \centering
    \setlength{\tabcolsep}{2mm}{
    \renewcommand\arraystretch{1.15}
    \begin{tabular}{c|c|c|c|c}
    \thickhline
    stage & output & layer name & PVT-T (\textbf{\ourmethod}) & PVT-S (\textbf{\ourmethod})\\
    \hline
    \multirow{4}*{res1} & \multirow{4}*{$56\times 56$} & Patch Embedding & $P_1=4; \ C_1=64$ & $P_1=4; \ C_1=64$\\
    \cline{3-5}
    & & Transformer & $\left[ \begin{array}{c} R_1=8\\N_1=1\\E_1=8 \end{array} \right ] (\textbf{\ourmethod}) \ \times 2$ & $\left[ \begin{array}{c} R_1=8\\N_1=1\\E_1=8 \end{array} \right ] (\textbf{\ourmethod}) \ \times 3$ \\
    \hline
    \multirow{4}*{res2} & \multirow{4}*{$28\times 28$} & Patch Embedding & $P_2=2; \ C_2=128$ & $P_2=2; \ C_2=128$\\
    \cline{3-5}
    & & Transformer & $\left[ \begin{array}{c} R_2=4\\N_2=2\\E_2=8 \end{array} \right ] (\textbf{\ourmethod}) \ \times 2$ & $\left[ \begin{array}{c} R_2=4\\N_2=2\\E_2=8 \end{array} \right ] (\textbf{\ourmethod}) \ \times 4$ \\
    \hline
    \multirow{4}*{res3} & \multirow{4}*{$14\times 14$} & Patch Embedding & $P_3=2; \ C_3=320$ & $P_3=2; \ C_3=320$\\
    \cline{3-5}
    & & Transformer & $\left[ \begin{array}{c} R_3=2\\N_3=5\\E_3=4 \end{array} \right ] (\textbf{\ourmethod}) \ \times 2$ & $\left[ \begin{array}{c} R_3=2\\N_3=5\\E_3=4 \end{array} \right ] (\textbf{\ourmethod}) \ \times 6$ \\
    \hline
    \multirow{4}*{res4} & \multirow{4}*{$7\times 7$} & Patch Embedding & $P_4=2; \ C_4=512$ & $P_4=2; \ C_4=512$\\
    \cline{3-5}
    & & Transformer & $\left[ \begin{array}{c} R_4=1\\N_4=8\\E_4=4 \end{array} \right ] (\textbf{\ourmethod}) \ \times 2$ & $\left[ \begin{array}{c} R_4=1\\N_4=8\\E_4=4 \end{array} \right ] (\textbf{\ourmethod}) \ \times 3$ \\

    \hline
    \thickhline
    \end{tabular}}
    \vspace{2pt}
    \caption{Architectures of PVT-based models with and without \ourmethod \ modules.}
    \label{Tab03}
\end{table*}

\begin{table*}[t]
    \centering
    \setlength{\tabcolsep}{2mm}{
    \renewcommand\arraystretch{1.15}
    \begin{tabular}{c|c|c|c}
    \thickhline
    stage & output & Swin-T (\textbf{\ourmethod}) & Swin-S (\textbf{\ourmethod})\\
    \hline
    \multirow{3}*{res1} & \multirow{3}*{$56\times 56$} & concat $4\times 4$, 96, LN & concat $4\times 4$, 96, LN\\
    \cline{3-4}
    && $\left[ \begin{array}{c} 7\times 7 {\rm \ window}\\{\rm dim} \ 96, {\rm head} \ 3\end{array} \right ] (\textbf{\ourmethod}) \ \times 2$ & $\left[ \begin{array}{c} 7\times 7 {\rm \ window}\\{\rm dim} \ 96, {\rm head} \ 3\end{array} \right ] (\textbf{\ourmethod}) \ \times 2$ \\
    \hline
    \multirow{3}*{res2} & \multirow{3}*{$28\times 28$} & concat $2\times 2$, 192, LN & concat $4\times 4$, 192, LN\\
    \cline{3-4}
    && $\left[ \begin{array}{c} 7\times 7 {\rm \ window}\\{\rm dim} \ 192, {\rm head} \ 6\end{array} \right ] (\textbf{\ourmethod}) \ \times 2$ & $\left[ \begin{array}{c} 7\times 7 {\rm \ window}\\{\rm dim} \ 192, {\rm head} \ 6\end{array} \right ] (\textbf{\ourmethod}) \ \times 2$ \\
    \hline
    \multirow{3}*{res3} & \multirow{3}*{$14\times 14$} & concat $2\times 2$, 384, LN & concat $4\times 4$, 384, LN\\
    \cline{3-4}
    && $\left[ \begin{array}{c} 7\times 7 {\rm \ window}\\{\rm dim} \ 384, {\rm head} \ 12\end{array} \right ] (\textbf{\ourmethod}) \ \times 6$ & $\left[ \begin{array}{c} 7\times 7 {\rm \ window}\\{\rm dim} \ 384, {\rm head} \ 12\end{array} \right ] (\textbf{\ourmethod}) \ \times 18$ \\
    \hline
    \multirow{3}*{res4} & \multirow{3}*{$7\times 7$} & concat $2\times 2$, 768, LN & concat $4\times 4$, 768, LN\\
    \cline{3-4}
    && $\left[ \begin{array}{c} 7\times 7 {\rm \ window}\\{\rm dim} \ 768, {\rm head} \ 24\end{array} \right ] (\textbf{\ourmethod}) \ \times 2$ & $\left[ \begin{array}{c} 7\times 7 {\rm \ window}\\{\rm dim} \ 768, {\rm head} \ 24\end{array} \right ] (\textbf{\ourmethod}) \ \times 2$ \\
    \hline
    \thickhline
    \end{tabular}}
    \vspace{2pt}
    \caption{Architectures of Swin-based models with and without \ourmethod \ modules.}
    \label{Tab04}
\end{table*}
\end{document}